%% file: arxiv_CameraReady.tex
\documentclass[10pt,twocolumn,letterpaper]{article}

\usepackage[pagenumbers]{cvpr} 

\def\cvprPaperID{8129} 
\def\confName{CVPR}
\def\confYear{2022}
 
\input{content/macros}

\begin{document}
\newcommand{\ShortName}{LEGO-Net\xspace}

\title{\ShortName: Learning Regular Rearrangements of Objects in Rooms}

\author{Qiuhong Anna Wei$^1$ \qquad Sijie Ding\thanks{Core contributions}\, $^1$ \qquad Jeong Joon Park\footnotemark[1]\, $^2$ \qquad Rahul Sajnani$^1$ \\ Adrien Poulenard$^2$ \qquad Srinath Sridhar$^1$ \qquad Leonidas Guibas$^2$\\\\
Brown University$^1$ \qquad Stanford University$^2$
}
\input{content/main/figures/teaser}
\input{content/main/text/01_abstract}

\input{content/main/text/02_intro}

\input{content/main/figures/method_overview.tex}
\input{content/main/text/03_related_works}

\input{content/main/text/04_method}

\input{content/main/figures/analysis_results.tex}
\input{content/main/figures/3dfront_results}
\input{content/main/text/05_results}

\input{content/main/text/06_conclusion}

{\small
\bibliographystyle{ieee_fullname}
\bibliography{egbib}
}
\input{supple_content.tex}

\end{document}

%% file: content/macros.tex
\usepackage{graphicx}
\usepackage{amsmath}
\usepackage{amssymb}
\usepackage{booktabs}
\usepackage{multirow}
\usepackage[pagebackref,breaklinks,colorlinks]{hyperref}
\usepackage[dvipsnames]{xcolor}
\usepackage{tabularx}

\definecolor{Gray}{gray}{0.85}
\definecolor{LightCyan}{rgb}{0.88,1,1}

\renewcommand{\paragraph}[1]{{\vspace{1mm}\noindent \bf #1}.}

\usepackage[capitalize]{cleveref}
\crefname{section}{Sec.}{Secs.}
\Crefname{section}{Section}{Sections}
\Crefname{table}{Table}{Tables}
\crefname{table}{Tab.}{Tabs.}

\newcommand{\todo}[1]{{\textcolor{red}{\textbf{#1}}}}

\usepackage{xspace}

\newenvironment{packed_itemize}
{\begin{itemize}
    \setlength{\itemsep}{1pt}
    \setlength{\parskip}{0pt}
    \setlength{\parsep}{0pt}
}{\end{itemize}}

\newcommand{\parahead}[1]{\noindent\textbf{#1}:\ }

%% file: content/main/figures/teaser.tex
\twocolumn[{%
\renewcommand\twocolumn[1][]{#1}%
\maketitle
\begin{center}
    \vspace{-1cm}
    \centering
    \captionsetup{type=figure}
    \includegraphics[width=0.88\textwidth]{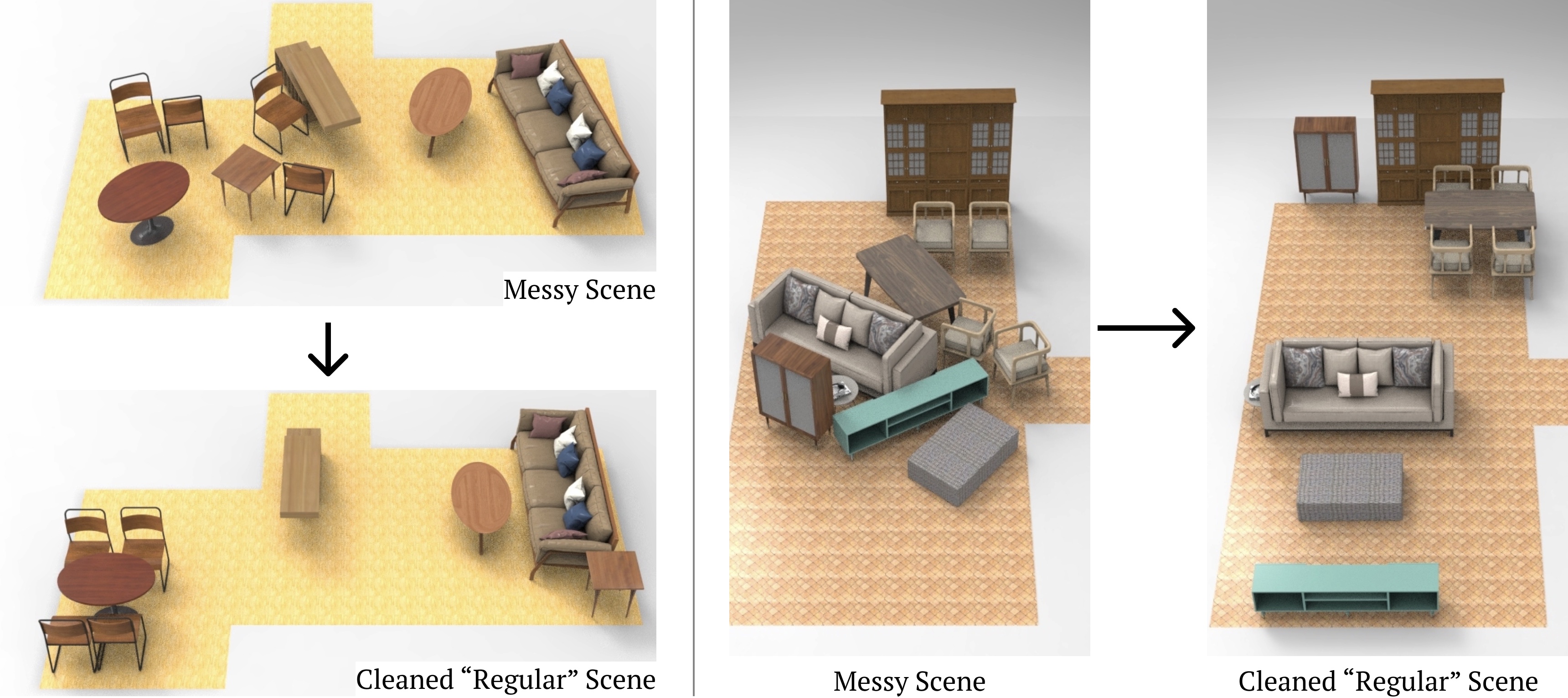}
    \captionof{figure}{
    \textbf{\ShortName} \textbf{\underline{LE}}arns to re\textbf{\underline{G}}ularly rearrange \textbf{\underline{O}}bjects in a messy indoor scene via an iterative denoising process.
    Different from scene synthesis or methods that require goal state specification, our method learns clean re-arrangements directly from data, retains the flavor of the original scene, and minimizes object travel distance.
    More about the project can be found at \href{https://ivl.cs.brown.edu/projects/lego-net}{ivl.cs.brown.edu/projects/lego-net}.
    \vspace{-0.4cm}
    }
    \label{fig:teaser}
\end{center}%
}]

%% file: content/main/text/01_abstract.tex
\begin{abstract}
%
{\let\thefootnote\relax\footnotetext{{$^*$Core contribution.}}}
\vspace{-0.2cm}
Humans universally dislike the task of cleaning up a messy room.
If machines were to help us with this task, they must understand human criteria for regular arrangements, such as several types of symmetry, co-linearity or co-circularity, spacing uniformity in linear or circular patterns, and further inter-object relationships that relate to style and functionality.
Previous approaches for this task relied on human input to explicitly specify goal state, or synthesized scenes from scratch -- but such methods do not address the rearrangement of existing messy scenes without providing a goal state.
In this paper, we present \ShortName, a data-driven transformer-based iterative method for \textbf{\underline{LE}}arning re\textbf{\underline{G}}ular rearrangement of \textbf{\underline{O}}bjects in messy rooms.
\ShortName is partly inspired by diffusion models -- it starts with an initial messy state and iteratively ``de-noises'' the position and orientation of objects to a regular state while reducing distance traveled.
Given randomly perturbed object positions and orientations in an existing dataset of professionally-arranged scenes, our method is trained to recover a regular re-arrangement.
Results demonstrate that our method is able to reliably rearrange room scenes and outperform other methods. We additionally propose a metric for evaluating regularity in room arrangements using number-theoretic machinery.
\vspace{-0.2cm}
\end{abstract}

%% file: content/main/text/02_intro.tex
\vspace{-1.0em}
\section{Introduction}
\label{sec:intro}
What makes the arrangement of furniture and objects in a room appear regular?
While exact preferences may vary, humans have by-and-large universally shared criteria of regular room arrangements: for instance, heavy cabinets are arranged to align with walls, chairs are positioned evenly around a table in linear or circular configurations, or night stands are placed symmetrically on the two sides of a bed. 
Humans also share a common dislike of physically performing the task of rearranging a messy room.
To build automated robotic systems that can guide or actually rearrange objects in a room, we first need methods that understand the shared human criteria for regular room rearrangements and respect the physical constraints of rearrangements.

Human criteria for regular rearrangements can be subtle and complex, including geometric rules of reflexional, translational, or rotational symmetry, linear or circular alignments, and spacing uniformity.
Functional and stylistic inter-object relationships are also important: for example, a TV tends to be in front of and facing a sofa, chairs are next to a table, etc.
Many of these criteria interact and, at times, conflict with one another.
As a result, in general, there is more than one desirable clean arrangement for any given messy arrangement.
In our setting, we further desire that the clean rearrangement we create to be informed by the initial messy arrangement -- and not be entirely different -- for multiple reasons.
First, there may have been a particular clean arrangement that gave rise
to the messy one -- and it may be desirable to recover a similar arrangement.
Second, we want to minimize the motion of objects as much as possible to respect the physical constraints and effort involved -- especially the motion of big and heavy furniture. 
Unfortunately, extant methods fail to capture these criteria: methods for scene synthesis from scratch~\cite{qi2018human,zhou2019scenegraphnet,li2019grains,luo2020end,zhang2020deep,zhang2020fast} ignore the initial state of objects in a room, and rearrangement methods often require scene-specific human input in the form of a goal state~\cite{batra2020rearrangement,shome2020synchronized} or language description~\cite{liu2022structformer,shridhar2022cliport}.

In this paper, we present \ShortName, a method for \textbf{\underline{LE}}arning re\textbf{\underline{G}}ular rearrangement of \textbf{\underline{O}}bjects in rooms directly from data.
Different from work that focuses on arranging new objects from scratch or requires goal state specification, we focus on \textbf{\underline{re}arranging} existing objects without any additional input at inference time.
We take as input the position, orientation, class label, and extents of room objects in a specific arrangement, and output a room with the same objects but regularly re-arranged.
\ShortName uses a transformer-based architecture~\cite{vaswani2017attention} that is, in part, motivated by recent denoising diffusion probabilistic models that learn a reverse diffusion process for generative modeling~\cite{song2020score,ho2020denoising,song2019generative}.
We learn human criteria for regular rearrangements from a dataset of professionally designed \emph{clean} (regular) scenes~\cite{fu20213d}, and represent each scene as a collection of objects and a floor plan.
Prior to training, we perturb the regular scenes to generate noisy configurations.
During training, our transformer learns to predict the original, de-noised arrangement from the perturbed scene and its floor plan.
During inference, instead of directly re-arranging scenes with our model, which would amount to na\"ive regression, we run a Langevin dynamics-like reverse process to iteratively denoise object positions and orientations.
This iterative process retains the flavor of original room state, while limiting object movement during re-arrangement.

We conduct extensive experiments on public datasets to show that our approach realistically rearranges noisy scene arrangements, while respecting initial object positions.
We also demonstrate that our method is able to generalize to previously unseen collection of objects in a wide variety of floor plans. 
Furthermore, we include extensive experimental results (e.g., \cref{fig:teaser} and \cref{fig: analysis}), including a new metric to evaluate regularity of re-arrangements, aimed at measuring the presence of sparse linear integer relationships among object positions in the final state (using the PSLQ algorithm~\cite{Ferguson1991}).
To sum up, we contribute:
\vspace{-0.2em}
\begin{packed_itemize}
\item A generalizable, data-driven method that learns to \textbf{regularly re-arrange} the position and orientation of objects in various kinds of messy rooms.
\item An iterative approach to re-arrangement at inference time that retains flavor of the original arrangement and minimizes object travel distance.
\item {An in-depth analysis of the performance and characteristics of the denoising-based scene rearrangement.}
\item {A new metric to measure the regularity of object arrangements based on integer relation algorithms.}
\end{packed_itemize}
\vspace{-1em}




%% file: content/main/figures/method_overview.tex
\begin{figure*}[h!]
    \centering
    \includegraphics[width= 1.0\textwidth]{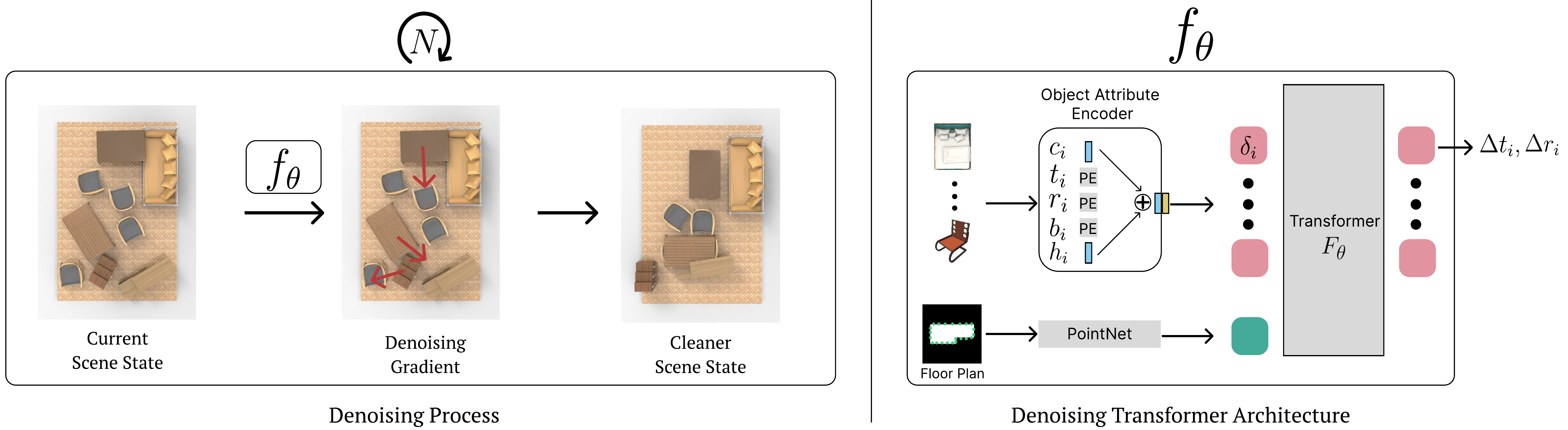}
    \vspace{-0.5cm} \caption{
    Pipeline overview. 
    LEGO-Net takes an input messy scene and attempts to clean it via iterative denoising. Given the current scene state, it computes the denoising gradient towards the clean manifold, and changes the scene accordingly. This denoising step is repeated until the scene is ``regular." On the right, we show our backbone transformer block $f_\theta$ that computes the denoising gradient at each step. It takes the scene attributes of the current state and outputs 2D transformations of each object that would make the scene ``cleaner".
    }
    \label{fig:architecture}
    \vspace{-1em}
\end{figure*}

%% file: content/main/text/03_related_works.tex
\section{Related Work}
\label{sec:relwork}
In this section, we discuss literature in two related areas: (1)~scene synthesis from scratch, (2)~scene rearrangement where an end goal is specified, and (3)~diffusion models.

\parahead{Indoor 3D Scene Synthesis}
Indoor room synthesis is the problem of synthesizing the layout of objects in a scene from scratch.
Many classical methods in the computer graphics literature use heuristics and guidelines to constrain the location of pre-specified objects~\cite{bukowski1995object,xu2002constraint,weiss2018fast,yu2011make}.
\cite{merrell2011interactive} identified a collection of functional, visual, and design constraints and formulated an optimization problem.
Work has also focused exclusively on inter-object relationships~\cite{li2021ifr}.
Other methods~\cite{yeh2012synthesizing} address the open world layout problem when objects are not pre-specified.

An alternative approach is to adopt procedural modeling using generative grammars~\cite{qi2018human,talton2011metropolis,purkait2020sg,devaranjan2020meta,chaudhuri2020learning,bokeloh2012algebraic}.
Some methods adopt the scenegraph representation and formulate it as a graph problem~\cite{zhou2019scenegraphnet,li2019grains,luo2020end,zhang2020deep,zhang2020fast,wang2019planit,para2021generative}.
Both procedural and graph-based methods often rely on curated data~\cite{fisher2012example}.
Some methods learn directly from data using neural networks, for instance from images~\cite{ritchie2019fast}.
Both SceneFormer~\cite{wang2021sceneformer} and ATISS~\cite{Paschalidou2021NEURIPS} introduce autoregressive methods for scene generation.
Different from all these methods, we focus on \emph{rearranging} rooms given an initial messy state.


\parahead{Scene Rearrangement}
Scene rearrangement takes an initial state of the scene and aims to bring it to a goal state specified by the user.
This task is deeply connected to planning in robotics~\cite{batra2020rearrangement,russell2010artificial,simon1962computer}.
Some works consider robot pushing and manipulation for rearrangement~\cite{ben1998practical,cosgun2011push,danielczuk2019mechanical,dogar2014object,king2016rearrangement,king2017unobservable,krontiris2014rearranging,scholz2010combining,shome2020synchronized,labbe2020monte}.
Many of these methods require datasets for training and often use datasets like AI2-THOR~\cite{kolve2017ai2}, Habitat~\cite{szot2021habitat} or Gibson~\cite{li2021igibson}.
Some of these methods operate on visual observations~\cite{king2016rearrangement}, while others assume fully-observed synthetic environments~\cite{cosgun2011push}.
To specify the goal state, some recent methods use language input~\cite{liu2022structformer,shridhar2022cliport} driven by large language models~\cite{radford2021learning,devlin2018bert}. Related to these advances in robotics, there have also been attempts to apply these specifically for room rearrangements~\cite{wang2020scenem,weihs2021visual}.

In this paper, we focus on the task of room rearrangements without the need to specify the goal state.
We directly learn arrangements that satisfy human criteria from professionally arranged dataset provided by 3D-FRONT~\cite{fu20213d}. Note that a concurrent work ~\cite{wu2022targf} addresses the same problem but with a focus on physical simulation, incorporating reinforcement learning and path planning.

\parahead{Denoising Diffusion Models}
2D Diffusion models \cite{song2020score,ho2020denoising,song2019generative} have emerged as a powerful technique for unconditional image synthesis, outperforming existing 2D generative models \cite{karras2020training, esser2021taming}. Diffusion  models have also seen great success in conditional image generations, receiving conditions in the form of class labels \cite{dhariwal2021diffusion}, text \cite{nichol2021glide,rombach2022high}, or input images \cite{saharia2022palette}. Various methods \cite{saharia2022palette,lugmayr2022repaint,meng2021sdedit} apply diffusion models for {\em restoring} corrupted or user-provided images to realistic images. Our method shares the same philosophy and adopts related techniques from the diffusion models, e.g., Langevin Dynamics, to project messy object configurations onto the manifold of ``clean" scenes.


%% file: content/main/text/04_method.tex
\vspace*{-0.08cm}
\section{Method}
\label{sec:method}
\vspace*{-0.01cm}
\subsection{Preliminaries}

Our method takes the position, orientation, class label, and extents of objects in a `messy' room as input and outputs a rearranged version in a `regular' state.
Since objects in rooms primarily move on the floor, we only consider 2D object pose, but our method can be combined with existing instance segmentation~\cite{wang2018sgpn,yi2019gspn} and canonicalization methods~\cite{sajnani2022condor} to directly operate from a 3D mesh or point cloud.
We represent each scene $X$ as an unordered set of $n$ objects and their attributes:
\vspace*{-0.05cm}
\begin{equation} \label{eq: object}
    X=\{ o_1,...,o_n \},\quad o_i=(c_i, t_i, r_i,b_i, h_i),
\end{equation}
where $c_i\in\mathbb{R}^{k}$, $t_i\in \mathbb{R}^2$, $r_i\in SO(2)$, and $b_i\in \mathbb{R}^2$ respectively denote the semantic class, translation, rotation, and bounding box dimensions of the object $o_i$. 
$h_i\in \mathbb{R}^{128}$ is the pose-canonicalized shape features obtained by running ConDor~\cite{sajnani2022condor} on each object's point cloud (see supplementary for details).
The furniture semantic class labels $c_i$'s are represented as one-hot vectors of the $k$ classes.
We represent the rotation $r_i\in SO(2)$  by the first column vector $[\cos(\theta),\sin(\theta)]^\intercal$ of its rotation matrix, following~\cite{zhou2019continuity} to represent $SO(2)$ without any discontinuity. 
Note that $t_i$ is normalized to be in $[-1,1]$ to have the same range as $r_i$'s sinusoidal representation to balance their importance during training.
We define that a scene $X^a$ is a rearrangement of $X^b$ (denoted $X^a\sim X^b$) iff there exists a bijection $\rho$ between object indices such that $h_i^a \approx h_{\rho(i)}^b$. 

\vspace*{-0.05cm}
\subsection{\ShortName: Learning Regular Room Rearrangements}
\cref{fig:architecture} shows our approach to solving the regular room rearrangement task.
Our method takes an input `messy' scene $\tilde{X}$ and outputs a rearranged, `regular' scene $X$. Towards this goal, we design a denoising Transformer~\cite{vaswani2017attention} $f_\theta$ that is trained to predict a clean scene given its perturbed version. During inference, we take an iterative approach as it gives us rich control of the rearrangement process, \eg,~moving lighter objects farther. 
At each time step $\tau$ of the denoising process, we pass the current scene state $\tilde{X}_\tau$ to the denoising Transformer $f_\theta$ that provides gradients towards the manifold of `clean' scenes.  We repeat the denoising process until the magnitude of the predicted gradient is small enough to finally obtain a clean manifold projection of the input scene.

\vspace*{0.05cm} \parahead{Manifold Projection via Denoising Autoencoder}
We now describe our approach from a manifold learning perspective. We can consider the input to our method as an off-manifold point $\tilde{X}$ (\ie,~a messy scene) and aim to project it to the closest point $X$ on the manifold of  `regular' scenes.
Our objective is to learn a function $f_\theta(\tilde{X})$ (with network parameters $\theta$) that finds such manifold projected point $X$, \ie, $f_\theta(\tilde{X}) \approx X.$ Motivated by the denoising autoencoders~\cite{vincent2008extracting} and their recent extensions to score-matching models~\cite{song2019generative,ho2020denoising}, we train such  $f(\tilde{X})$ by perturbing the regular data $X$ employing a noise kernel $q_\sigma(\tilde{X}|X)$ with noise parameter $\sigma$. The training is done by minimizing a denoising objective function:
\vspace*{-0.1cm}\begin{equation}\label{eq:denoising}
\mathcal{E}_{dn}(\theta)=\mathbb{E}_{q_\sigma(\tilde{X},X)}\left[ \mathcal{L}_{\text{dn}}\left(f_\theta(\tilde{X}),X\right) \right].
\end{equation}
The joint distribution $q_\sigma(\tilde{X},X)=n(\sigma)q_\sigma(\tilde{X}|X)q_0({X}),$ where $n(\sigma)$ is the distribution of the noise parameter and $q_0({X})$ is the discrete uniform distribution of the training examples. Here, the loss $\mathcal{L}_{dn}$ is defined as the average distance between the pairs of objects in $f_\theta(\tilde{X})$ and $X$:
\vspace*{-0.1cm} \begin{align}
    \mathcal{L}_{\text{dn}}=\frac{1}{n}\sum_{i=1}^{n}||\tilde{t_i}-t_i||_2^2+||\tilde{r_i}-r_i||_2^2\nonumber\\
    +\lambda_1(||\tilde{t_i}-t_i||_1+||\tilde{r_i}-r_i||_1),
\end{align}
where $\tilde{t}_i$ and $\tilde{r}_i$'s are the object rotation and translation parameters of $f_\theta(\tilde{X})$, and $\lambda_1$ is a balancing parameter for the L1 loss. While we can use the object correspondences from the perturbation process, we choose to re-establish object pairing by computing Earth Mover's Distance~\cite{rubner2000earth} between the same class of objects in original and perturbed scenes.

Intuitively, the network $f_\theta(\tilde{X})$ learns to project $\tilde{X}$ to the clean manifold. However, it is not trained to find a random point in the clean manifold, but rather tries to find $X$ that shares similarities with $\tilde{X}$, depending on the noise level.

\vspace*{0.1cm} \parahead{Connection to Score-based Models}
While the trained denoising network $f_\theta$ can theoretically be applied to clean a messy scene directly, in practice, the quality of the output is suboptimal, as shown in \cref{sec:experiments}.
This is because the network $f_\theta$ is trained with a {\em regression} loss, which is known to fit to the average state of the conditional distribution $q_\theta(X|\tilde{X})$, leading to blurry predictions \cite{zhang2016colorful,isola2017image}.

Recently, score-based generative models (and the closely related diffusion models) \cite{ho2020denoising,song2019generative} have shown impressive image generation results using a trained denoiser.   
The score-based approaches \cite{vincent2011connection,song2019generative} approximate the gradient of likelihood of the perturbed data distribution $q_\sigma(\tilde{X})$ with a neural network $s_\phi$,   and showed that the optimal network $s^*_\phi$ for the denoising objective  
$
    \mathbb{E}_{q_\sigma(\tilde{X},X)}\left[ \left\lVert s_\phi(\tilde{X})- \nabla_{\tilde{X}} \log q_\sigma(\tilde{X}|X) \right\rVert^2 \right]
$
satisfies $s^*_\phi(\tilde{X})\approx \nabla_{\tilde{X}} \log q_\sigma(\tilde{X}).$ Assuming a zero-mean Gaussian noise kernel $q_\sigma(\tilde{X}|X)$, the score-based network training objective becomes:
\vspace*{-0.12cm} \begin{equation}\label{eq:score}
    \mathcal{L}_{score}(\phi)=\mathbb{E}_{q_\sigma(\tilde{X},X)}\left[ \left\lVert s_\phi(\tilde{X})- \frac{X-\tilde{X}}{\sigma^2} \right\rVert^2 \right].
\end{equation} \vspace*{-0.1cm}

Since the trained score network $s^*_\phi$ approximates the gradient of the data distribution, it can be used for autoregressively optimizing noisy data onto the manifold of clean data. In our context, $X-\tilde{X}$ amounts to the difference in object transformations between the clean and perturbed scenes, and the direction $\frac{X-\tilde{X}}{\sigma^2}$ predicted by $s^*_\phi(\tilde{X})$ is clearly towards the clean scene manifold.

\input{content/main/figures/process.tex}

\vspace*{0.08cm} \parahead{Rearrangement with Langevin Dynamics}
After training with Eq.~\ref{eq:denoising}, we have a function that is optimized to approximate the denoised projection of an input, \ie, $f_\theta^*(\tilde{X})\approx X.$ Given that $s^*_\phi(\tilde{X})\approx \frac{X-\tilde{X}}{\sigma^2},$ we have that $s^*_\phi(\tilde{X})\propto f_\theta^*(\tilde{X})-\tilde{X}.$ We then follow the score-based methods to adopt Langevin dynamics \cite{song2019generative,ho2020denoising} to recursively denoise the scene (see \cref{fig:process}) using the estimated gradients: \vspace*{-0.08cm} \begin{equation}\label{eq:langevin}
    \tilde{X}_{\tau+1}=\tilde{X}_{\tau}+\alpha(\tau)\left(f_\theta^*(\tilde{X}_\tau)-\tilde{X}_\tau\right)+\beta(\tau)z_\tau,
\end{equation} \vspace*{-0.3cm} 

\noindent where $\alpha(\tau)$ and $\beta(\tau)$ are monotonically-decreasing functions of time $\tau$ that is heuristically designed to balance the Langevin dynamics and $z_\tau \sim \mathcal{N}(0,\,1)$. 
We run the recursive computation until the magnitude of the gradient is small enough (\ie, $\lVert f_\theta^*(\tilde{X}_{\tau_i})-\tilde{X}_{\tau_i}\rVert<\kappa$, for constant $\kappa$) for $k$ consecutive iterations for some constant $k$.

\subsection{Architecture}
\ShortName follows the recent success in the scene synthesis community to adopt the Transformer architecture~\cite{vaswani2017attention} to represent our denoising function $f_\theta$, as illustrated in Fig.~\ref{fig:architecture}.
Given an input scene $\tilde{X}$, a Transformer encoder network $F_\theta$ takes in $|\tilde{X}|+1$ number of 512-dimensional tokens $\delta_i$'s corresponding to the objects in the scene, as well as the room floor plan. Then, the network outputs absolute translation and rotation predictions for all object tokens (excluding the layout token), \ie,~$F_\theta:\mathbb{R}^{(|\tilde{X}|+1)\times 512} \mapsto \mathbb{R}^{|\tilde{X}|\times 4}$. We then apply the outputs to guide the translation and rotation of each object in $\tilde{X}$. The denoiser $f_\theta$ is defined to include both operations. Therefore, the final processed scene is a rearrangement of the input scene: $f_\theta(\tilde{X})\sim \tilde{X}$.

\parahead{Input Object Attribute Encoding}
We use the following process to abstract $o_i$ into a token vector $\delta_i$. We employ positional encodings of 32 frequencies, and an additional linear layer for $r_i$, to independently process $t_i,r_i,b_i$ into vectors in $\mathbb{R}^{128}$.
For object class $c_i$, we employ a $2$-layer MLP with leaky ReLU activation to process the one-hot encoding into an attribute in $\mathbb{R}^{128}$. Finally, we optionally process a pose-invariant shape feature $h_i$ from ConDor \cite{sajnani2022condor} with a $2$-layer MLP to obtain a feature in $\mathbb{R}^{128}$. The above attribute features are then concatenated and processed with a $2$-layer MLP to form an object token $\delta_{i}\in \mathbb{R}^{512}.$ We refer readers to supplementary for the full details of the processing.

\parahead{Floor Plan Encoder}
Floor plans designating room boundaries both impose important realistic constraints and provide regularity information for the scene rearrangement task. Therefore, we pass the room layout in the form of an object token to the transformer so that other objects can attend to it. We employ a floor plan encoder to tokenize the floor plans as follows.  
We uniformly sample 250 points from the contour of the floor plan. These points along with their 2D surface normals are then processed with a simplified version of PointNet \cite{qi2017pointnet}. 
Finally, we specifically assign one bit of the $512$ transformer input dimensions (for both objects and floor plans) to distinguish floor plan `objects' from normal `objects'. The final output of the floor plan encoder for each scene is a feature in $\mathbb{R}^{512}$.

\parahead{Transformer Architecture}
We use our custom positional encodings and procedures to prepare the tokens but use the original Transformer encoder architecture without notable modifications. We use $8$ multi-headed attentions with $512$-dimensional hidden layers and $512$-dimensional key, query, and value vectors. The output of the transformer network is the estimated object transformations, a ${|X|\times 4}$ matrix.


\subsection{Training and Inference}
\vspace*{-0.1cm}\parahead{Data}
We employ the 3D-FRONT dataset \cite{fu20213d} for the task of indoor scene rearrangement. For each valid clean scene in the dataset, we preprocess it into $X = \{o_1, ..., o_n\}$ and extract the contour of its floor plan. 

\parahead{Training}
We use the denoising auto-encoder formulation of Eq.~\ref{eq:denoising} to train our denoiser function $f_\theta.$ We uniformly randomly sample training examples and sample a noise level $\sigma$ from a normal distribution. The sampled examples are perturbed using an independent Gaussian kernel with standard deviation $\sigma$. For the perturbation, we do not consider objects going outside of the floor plans or colliding with one another. Each perturbed scene uses its original clean scene as the source of ground truth but re-establishes object correspondence through Earth Mover's Distance assignment to enable invariance among identical objects and further promote distance minimization in movement prediction. We use Adam optimizer with a learning rate of $10^{-4}$ to train.

\parahead{Inference}
During inference, we use the Langevin Dynamics scheme of Eq.~\ref{eq:langevin} to iteratively project a messy scene input $\tilde{X}$ onto the manifold of clean scenes. We select a hyperbolic function $\alpha(\tau) = \alpha_0 / (1 + a_1*\tau)$ to regulate the step size. We additionally select an exponential $\beta(\tau)= \beta_0 * b_1^{\lfloor \tau/b_2 \rfloor}$, where effectually $\beta_0$ is multiplied by $b_1$ every $b_2$ iterations, to adjust the level of noise as the denoising process proceeds. Refer to supplementary for details.  



%% file: content/main/figures/process.tex
\begin{figure}
    \centering
    \vspace{-0em}
    \includegraphics[width= 0.95\linewidth]{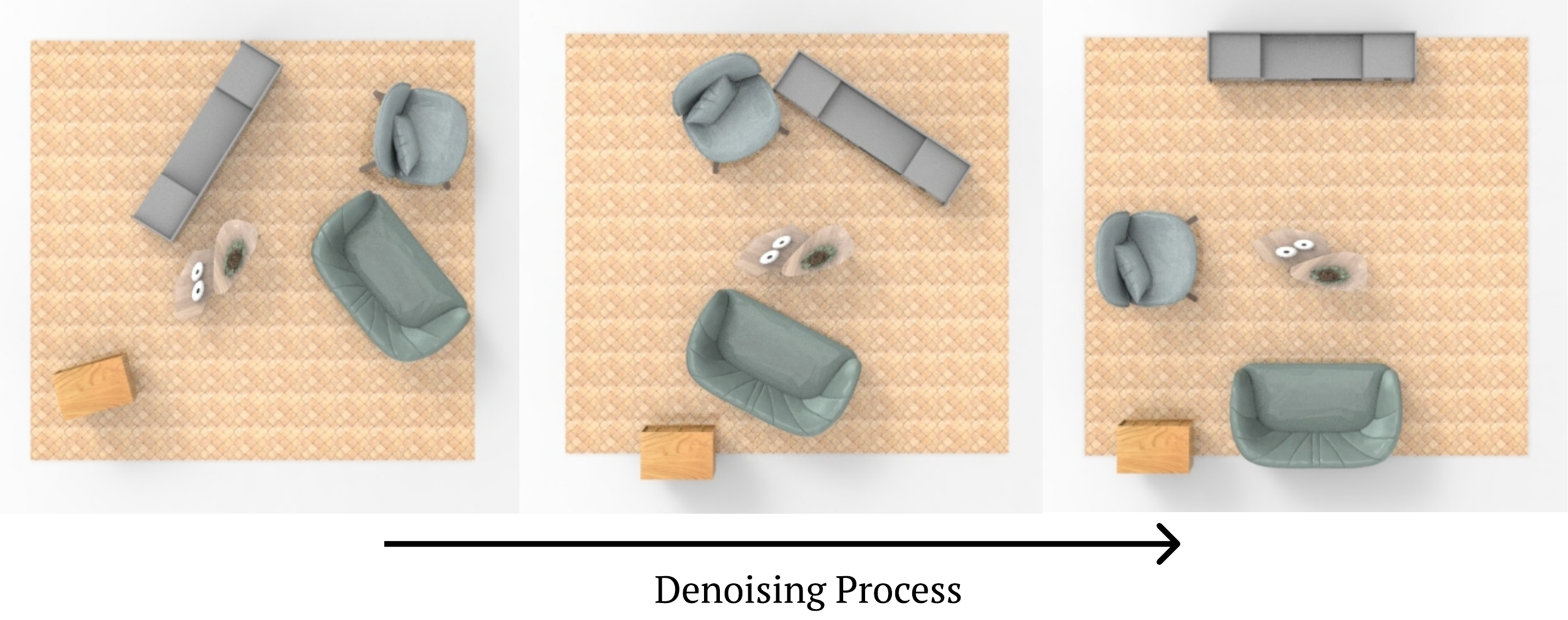}
    \caption{
    Instead of directly regressing the final rearranged state which can lead to non-diverse, suboptimal results, we adopt an iterative strategy based on Langevin Dynamics.
    At each step in our process (left to right), we gradually ``de-noise'' the scene until it reaches a regular state.
    During training, we follow the reverse process, \ie,~perturb clean scenes to messy state (right to left).
    }
    \label{fig:process}
    \vspace{-0.4cm}
\end{figure}

%% file: content/main/figures/analysis_results.tex
\begin{figure}
    \centering
    \includegraphics[width=1.0\linewidth]{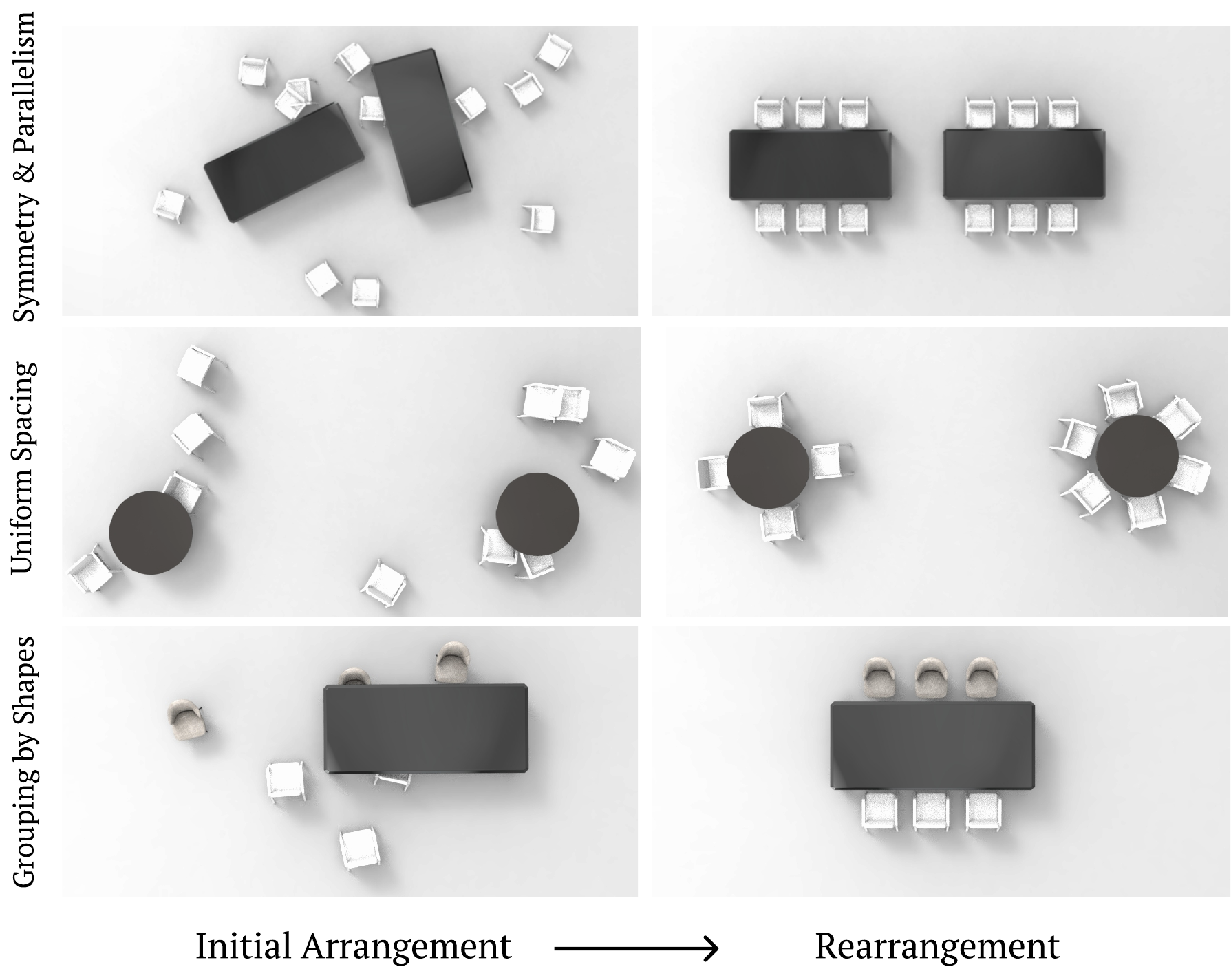}
    \vspace*{-0.3cm}\caption{Regularities learning results. We train our denoising network to learn three different regularities. LEGO-Net successfully learns the complex regularity rules as demonstrated by the iterative denoising results shown on the right.}
    \label{fig: analysis}
    \vspace{-0.4cm}
\end{figure}

%% file: content/main/figures/3dfront_results.tex


\begin{figure*}
    \centering
    \includegraphics[width= 0.99\textwidth]{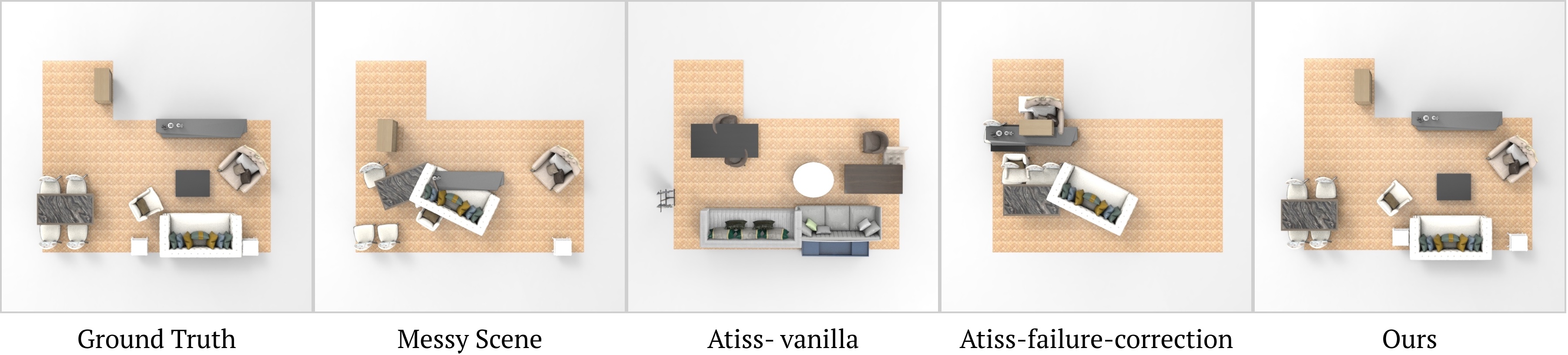}
    \caption{Comparison against ATISS \cite{Paschalidou2021NEURIPS} on 3D-FRONT dataset. As the state-of-the-art scene synthesis method, ATISS is able to produce a realistic scene (3rd column) given the floor plan, but the generated objects and arrangements are entirely different. On the other hand, we solve the problem of re-arranging the given messy scene, directly using the existing objects. While ATISS has shown failure correction technique that solves similar problem to ours, we observe that when the scene is highly noisy, their algorithm tends to deteriorate significantly. Moreover, being a one-shot prediction method, ATISS-failure does not consider the moving distance of the new arrangement. }
    \label{fig: 3dfront}
    \vspace{-3mm}
\end{figure*}

%% file: content/main/text/05_results.tex
\section{Experiments} \label{sec:experiments}
We conduct a number of experiments to test \ShortName's ability to automatically capture scene regularities from data.
To this end, we prepare two testbeds for experiments: our custom-designed \textbf{Table-Chair} environment where we mathematically constructed the regularities among objects and \textbf{3D-Front}~\cite{fu20213d}, which contains tens of thousands of synthetic rooms designed by professionals.
The Table-Chair dataset is useful because we can model one regularity at a time and quantify network performance.
3D-Front dataset exhibits complex and subtle rules that designers commonly perceive as ideal configurations, \eg,~geometry, semantic relations, styles, and functionalities.

\vspace*{-0.05cm}\subsection{Capturing Regularities from Data}
In the Table-Chair environment, we study four main regularities: symmetry, parallelism, uniform spacing, and grouping by shapes. For each of the proposed experiments, we generate clean scenes based on the designed rules. Then, for training, we perturb the scenes on the fly to generate clean-messy pairs and re-associate objects within each class through Earth Mover's Distance assignment to train a network with the loss of Eq.~\ref{eq:denoising}. We measure each task with the success rate of the rearrangement, whose specific criteria we discuss in the supplementary.

\parahead{Symmetry and Parallelism}
One of the most important notions of regular arrangement is symmetry, which involves both object-object and room-level symmetries.
We use a setup of 2 groups of rectangular tables and chairs. 
We vertically align the 2 tables and horizontally distribute them at a distance uniformly drawn from a fixed range. We arrange 3 chairs in a linear row on one side of the table and 3 chairs in another linear row on the opposite side.

\parahead{Uniform Spacing}
We prepare a highly-challenging setup to stress test \ShortName's ability to capture the concept of uniform spacing. In this setup, we have 2 circular tables, each with 2-6 chairs randomly and uniformly rotated around them. The network has to deal with the unknown number of chairs and the pair-wise spacing.

\parahead{Grouping by Shape}
We test \ShortName's ability to group objects based on their pose-invariant shapes. We augment our setup in `Symmetry and Parallelism' to include 2 types of chairs with different shapes. We arrange the scenes such that chairs with the same shapes are on the same side of the table. The pose-invariant shape features $h_i\in o_i$ from \cref{eq: object} provide the necessary shape information.

\parahead{Results}
We visualize the rearrangement results of \ShortName for the above three cases in Fig.~\ref{fig: analysis}.
Across the board, the denoising network successfully learns to capture these important regularities from data, without explicit supervision about the underlying rules.
The success rate of each task is shown in ~\cref{table:tablechair_metrics}.
As expected, directly applying the regression-trained network $f_\theta$ results in the worst results.
\vspace{-1mm}

\begin{table}[t]
\centering
\begin{tabular}{ c c c c }
\toprule
 & \vtop{\hbox{\strut Symmetry \&}\hbox{\strut Parallelism $\uparrow$}} & \vtop{\hbox{\strut Uniform}\hbox{\strut Spacing$\uparrow$}}& \vtop{\hbox{\strut Grouping}\hbox{\strut by Shape$\uparrow$}} \\ \hline
Direct & 16\% & 18.6\% & 23.6\% \\
Grad. w/o noise & 91.2\% & 96\% & 87.8\% \\ 
Grad. w/ noise & \textbf{91.4}\% & \textbf{97.2}\% &
\textbf{89.2}\% \\ 
\bottomrule
\end{tabular}
\caption{Denoising success rate by regularities and inference strategies. For the three regularities shown in Fig.~\ref{fig: analysis}, we measure the success rate of \ShortName for each of the inference variants.}
\vspace{-0.3cm}
\label{table:tablechair_metrics}
\end{table}

\subsection{3D-Front Experiments}
\vspace{-1mm}
We benchmark \ShortName's ability to conduct regular scene rearrangements on the bedrooms and livingrooms of the 3D-FRONT dataset, which respectively contains 2338/587 and 5668/224 scenes for train/test splits.

We train our \ShortName as described in \cref{sec:method} with $\sigma \sim \mathcal{N}(0, 0.1^2)$. While we maintain a single denoising network $f_\theta$, we explore three variants of inference algorithms to provide greater insight of our approach: (1) LEGO-Net  {\em direct}, (2) {\em grad. with noise}, and (3) {\em grad. w/o noise} respectively denote the inference strategy of predicting the clean outcome with one network pass, running Langevin Dynamics of \cref{eq:langevin} with noise term $\beta \neq 0$, and $\beta=0$.

\paragraph{Baselines} We compare our rearrangement results against the current SOTA scene synthesis method, ATISS \cite{Paschalidou2021NEURIPS}. While ATISS is designed to synthesize a scene from scratch rather than to rearrange one, it provides an auto-regressive generative model that can be flexibly applied to our task. Specifically, we use three variants of ATISS that share the same network weights. First, ATISS {\em vanilla} performs its original scene synthesis task given a floor plan. Second, ATISS with {\em  labels} performs object placement using a predefined set of objects per scene. Third, ATISS {\em failure-correction} takes a noisy scene and cleans it up by iteratively finding an object with low probability and re-placing it within the current scene. This variant of ATISS is given the same perturbed scene as \ShortName and aims to clean the scene.  Note that we omit to compare against prior works that have already been compared against ATISS, e.g., \cite{wang2019planit,wang2021sceneformer,ritchie2019fast}. We could not find a prior data-driven method that is designed to solve the same rearrangement problem as ours.

\input{content/main/tables/main_table.tex}

\paragraph{Metrics}
To gauge how well \ShortName captures datasets' regularities, we adopt the popular FID and KID scores. These metrics compare the closeness of statistics of two data distributions. We follow prior works \cite{wang2018deep,Paschalidou2021NEURIPS} to render ground truth and generated scene arrangements from top-down views and compute the metrics in the image space. Note that KID is more applicable to our setting because FID is known to present huge bias when the number of data is low. Another important criterion for our rearrangement task is how much distance the objects travel between the initial and final scene states. Similarly, when applicable, we measure the Earth Mover's Distance (EMD) between the ground truth scene and our cleaned-up scene.

Finally, we introduce a new metric that measures scene regularities by finding integer relations among object positional coordinates $t_i$'s. To do this, we select two or three random objects within a scene and check if we can find integral $a_i$'s that satisfy:
\begin{equation}
    a_1t_1+...+a_nt_n = 0, \quad 0<|a_i|<\eta\,, \forall a_i
\end{equation} 
where $\eta$ sets the maximum magnitude of the coefficients. Intuitively, these integer relations can capture regularities such as colinearities ($-t_1+t_2=0$) and symmetries ($t_1-2t_2+t_3=0$). See supplementary for detailed descriptions.


\paragraph{Results}
We conduct the 3D-FRONT arrangement experiments with five algorithms (three ATISS variants and two of ours) and compute their metrics. The main numerical results, which can be found in Tab.~\ref{tab:3dfront_main}, show that \ShortName outperforms all variants of ATISS, including the failure-correction variant that tackles the same object cleaning problem as demonstrated in the original paper. 

In Fig.~\ref{fig: integer_main}, we plot the chance of finding integer relations in scenes perturbed with different noise levels, which peaks for the original clean 3D-FRONT scenes and sharply decreases as noise is added. Also, note that the rearranged scenes of \ShortName demonstrate high regularities according to this measure, outperforming the results of ATISS variants. See supplementary for more experiment details.

\input{content/main/figures/main_integer.tex}

Qualitatively, as shown in Fig.~\ref{fig: 3dfront}, \ShortName is able to robustly project messy scenes onto clean manifolds. While ATISS and ATISS with {\em labels} were able to synthesize realistic rooms, their object arrangements are entirely different from the original input scene. Importantly, we notice that ATISS \emph{failure correction} leads to unexpectedly low-quality results. We hypothesize that this is due to their discrete, one-object-at-a-time strategy, which can easily fall into the local minimum of the likelihood space. In contrast, our score-based iterative denoising leads to robust success rates.

\input{content/main/figures/NoiseLevel.tex}

\subsection{Analysis}
To more deeply understand the behavior of our system, we analyze and discuss important aspects of \ShortName. We refer to supplementary for more analysis of our method.

\paragraph{Denoising Strategy}
As we discuss throughout Sec.~\ref{sec:experiments}, we explore three inference strategies, namely direct, gradient with noise, and gradient without noise. For the 3D-FRONT experiment, we report that the {\em grad. without noise} variant consistently outperforms the other variants. However, we believe that this is likely because we used relatively low noise to the scenes (std $0.1$) to  more naturally simulate messy indoor rooms. Indeed, our experiment on the synthetic environments (Tab.~\ref{table:tablechair_metrics}) with larger noise (std $0.25$) shows that the \emph{grad. with noise} variant outperforms. The results suggest that adding noise during Langevin dynamics allows a better success rate for highly noisy data, but at the cost of losing accuracy in recovering the originals (as shown in Tab.~\ref{tab:3dfront_main}).

\paragraph{Cleaning Uncertainty}
\ShortName is trained to handle input perturbations at various noise levels. In the high-noise regime, there is high uncertainty on the structure of the original information. As input noise increases, our denoising process converges into an unconditional generative model. 
On the other hand, \ShortName has the capacity to capture original regularities when the noise is low, leading to almost precise reconstruction  of the original scenes. We visually show these insights in  Fig.~\ref{fig:noiselevel}.

\paragraph{Out-of-Distribution Inputs}
We showcase \ShortName's ability to handle scenes perturbed with noise patterns significantly different from the one used in training, \ie,~zero-mean Gaussian. In the first example, we only perturb chairs. Secondly, we perturb the scene only in the translation dimensions without rotations. 
Shown in \todo{}Fig.~\ref{fig: special}, \ShortName can successfully handle out-of-distribution inputs, demonstrating the robustness and versatility of our algorithm. 

\input{content/main/figures/special.tex}

%% file: content/main/tables/main_table.tex
\begin{table*}[t]
 \renewcommand*{\arraystretch}{1.1}
    \centering
		\small
    \begin{tabular}{@{\hskip 1mm}c c c c c c |c c c c@{\hskip 1mm}}
				\toprule
					       & & \multicolumn{4}{c}{Living Room}                                                                                           & \multicolumn{4}{c}{Bedroom}                \\
					       & & KID $\!\downarrow$     & FID $\!\downarrow$   &\centering \vtop{\hbox{\strut Distance}\hbox{\strut \,\,Moved}} $\!\downarrow$   & \vtop{\hbox{\strut EMD }\hbox{\strut to GT}} $\!\downarrow$  & KID $\!\downarrow$     & FID $\!\downarrow$   &\centering \vtop{\hbox{\strut Distance}\hbox{\strut \,\,Moved}} $\!\downarrow$   & \vtop{\hbox{\strut EMD }\hbox{\strut to GT}} $\!\downarrow$  \\
        \midrule
        \multirow{3}{*}{ATISS \cite{Paschalidou2021NEURIPS}}&{\em vanilla}    & 96           &	44.55          & ---            	          & ---	          &   33                & 50.49                   & ---    & ---      \\
        &{\em labels}	    & 119          &	45.45        & ---          & 0.3758  	      & 49 	      & 52.62                 & ---                 & 0.5482          \\
        &{\em failure-correction} 	& 280          &	61.55	     & 0.1473  	     & 0.3378 		  & 240 		  & 73.95                 & 0.2025                 & 0.4673          \\  \midrule
        \multirow{2}{*}{LEGO-NET (ours)}&grad. w/ noise    & 67	        &	39.19          & 0.091          &  0.125		      & 37          & 49.76                  & 0.052                   & 0.086           \\
        
        &grad. w/o noise 		& \textbf{51}           & \textbf{37.47}            & \textbf{0.086} 	         & \textbf{0.117}	          & \textbf{27}		      & \textbf{48.43  }                & \textbf{0.0492}                   & \textbf{0.0815} \\
				\bottomrule
    \end{tabular}
    \captionsetup{width=1.0\textwidth}
    \caption{Quantitative experiment results using KID $\times10,000$, FID, distance moved, and Earth Mover's Distance (EMD) against the ground truth arrangements. All scenes are situated within $[-1,1]^2$ canvas. Note that ATISS {\em vanilla} and ATISS {\em labels} start from empty floor plans and thus the distance moved metric is not applicable. ATISS {\em failure-correction} takes a noisy scene and iteratively resamples low-probable objects, and is thus directly comparable to our method.}
    \label{tab:3dfront_main}
\end{table*}

%% file: content/main/figures/main_integer.tex
\begin{figure}
    \centering
    \includegraphics[width= 1.0\linewidth]{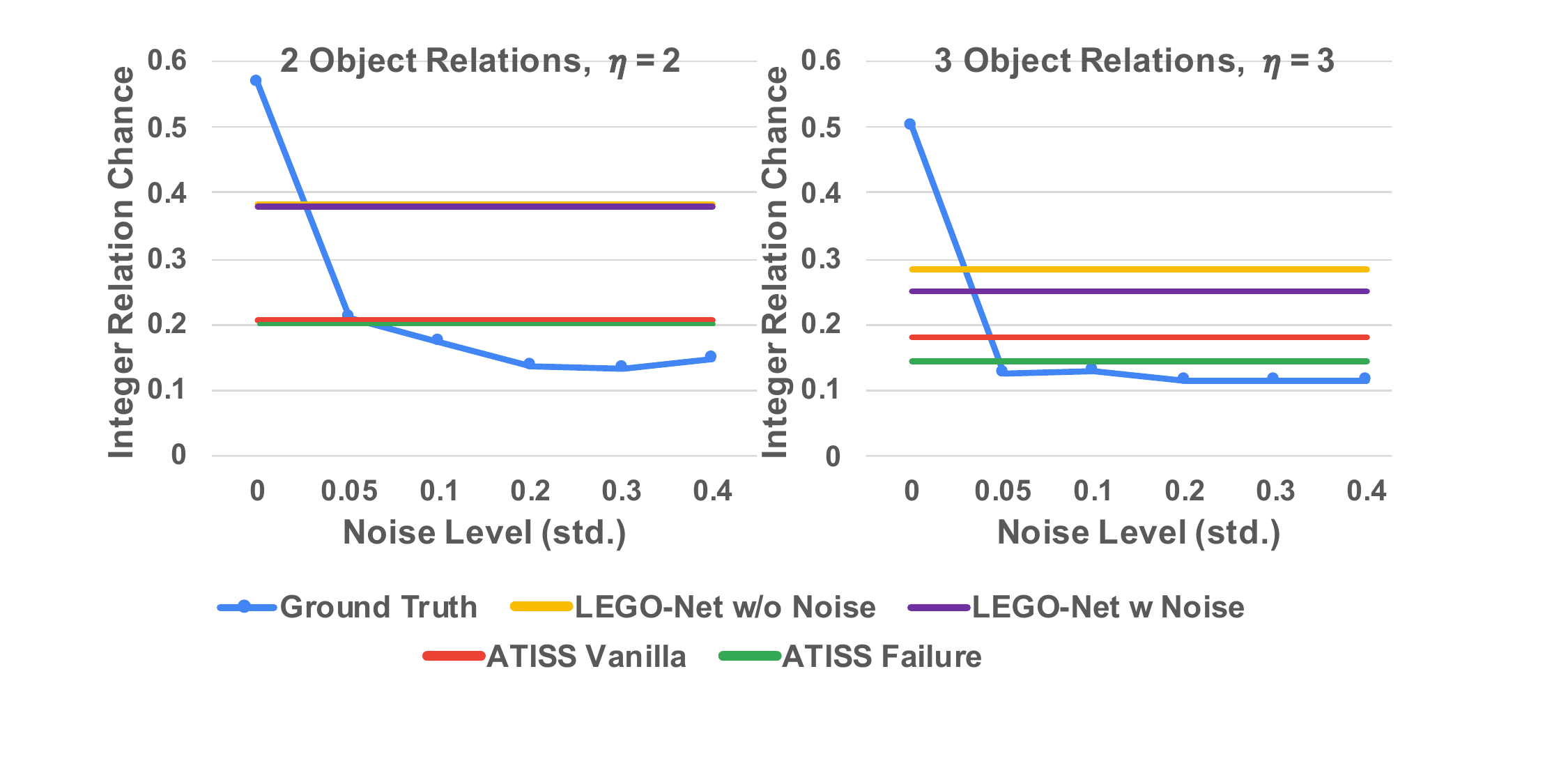}
    \vspace{-0.5cm}
    \caption{Integer relation occurences. We measure the chance of finding integer relations between the coordinates of two (left) and three (right) objects within a Living Room scene. Perturbing ground truth scenes sharply decreases the integer relation occurences, showing they are useful metric of regularities. Note that LEGO-Net outperforms ATISS variants in this metric.
    }\label{fig: integer_main}
    \vspace{-0.3cm}
\end{figure}

%% file: content/main/figures/NoiseLevel.tex
\begin{figure*}[h!]
    \centering
    \vspace{-0.1em}
    \includegraphics[width= 0.99\textwidth]{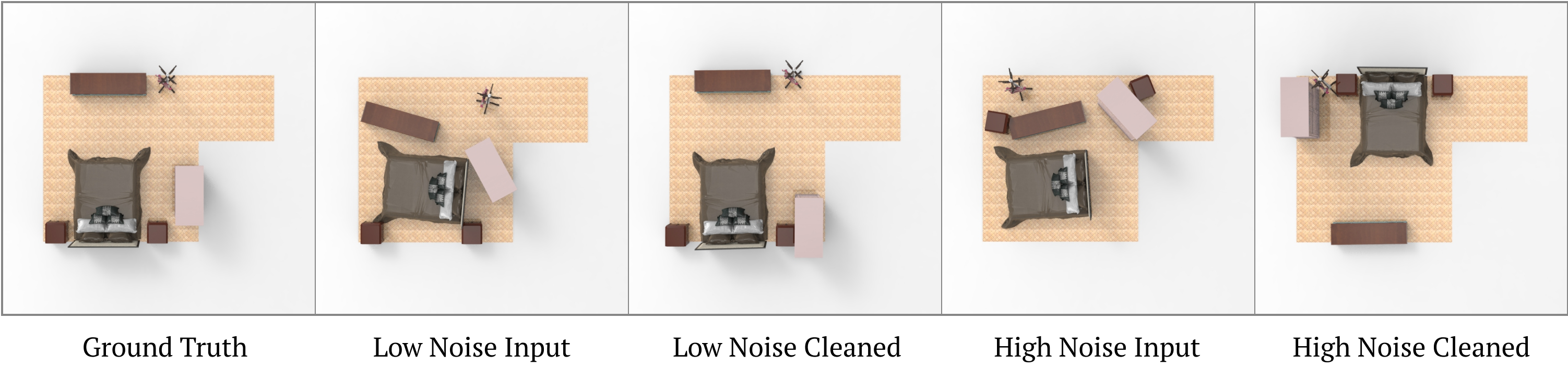}
    \caption{LEGO-Net denoising results on different noise levels. When the perturbation added to the scene is low, LEGO-Net is able to closely reconstruct the clean version of the scene. In contrast, when the noise level is high, our denoising process finds a different realization of a regular scene, behaving more like an unconditional model. Similar phonemena have been observed by 2D diffusion projects, e.g., SDEdit \cite{meng2021sdedit}.
    }
    \label{fig:noiselevel}
    \vspace{-0.3cm}
\end{figure*}

%% file: content/main/figures/special.tex
\begin{figure}[h!]
    \centering
    \includegraphics[width=1.0\linewidth]{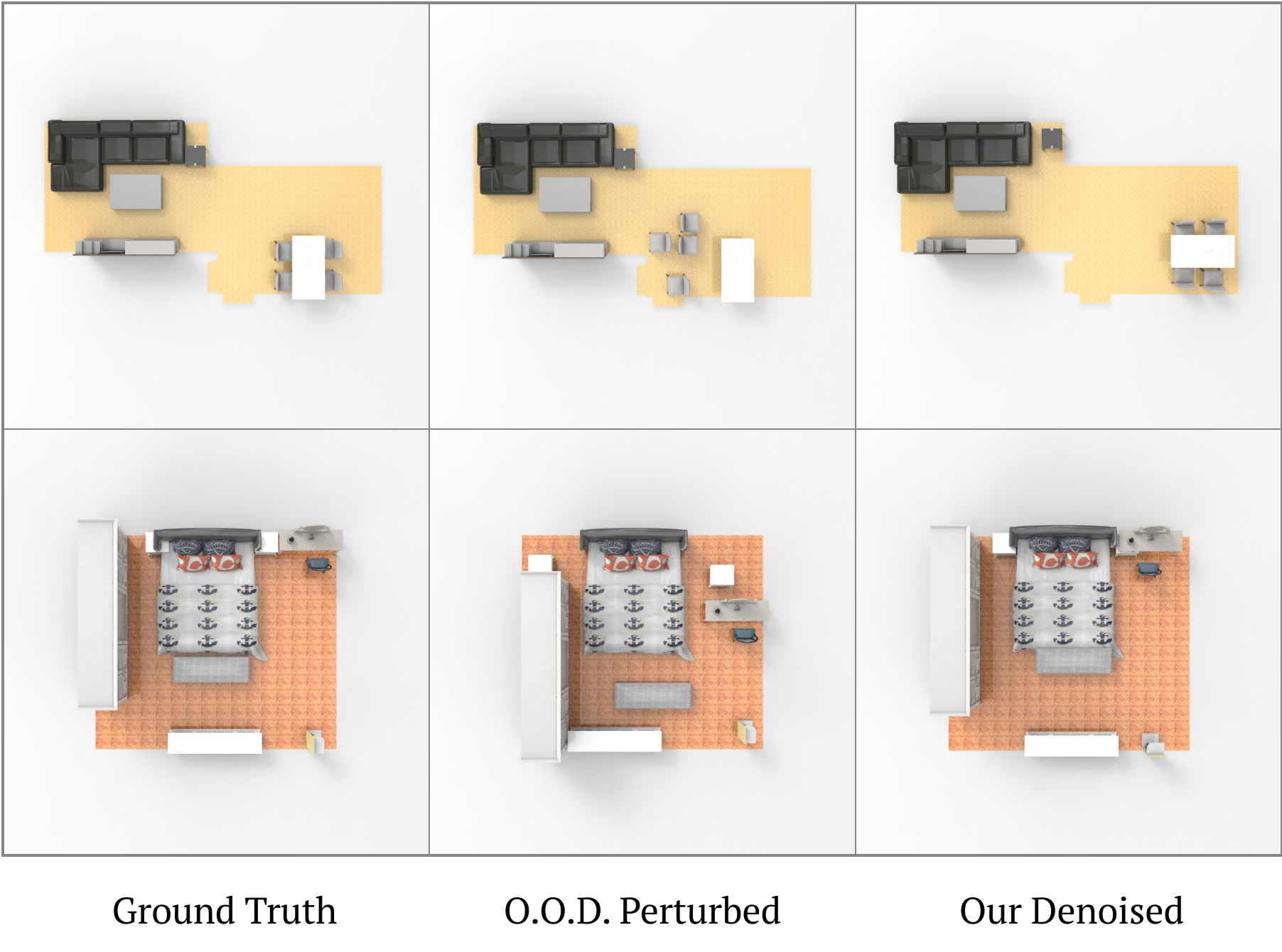}
    \vspace{-1em}
    \caption{Out-of-distribution test. While our model is trained to denoise Gaussian noise, it demonstrates strong robustness to out-of-distribution inputs. In the first row, only the chairs are perturbed. In the second row, we perturbe the scene with translation noise only. Zoom-in for details.}
    \label{fig: special}
    \vspace{-0.5cm}
\end{figure}

%% file: content/main/text/06_conclusion.tex
\section{Conclusion}
In this paper, we presented \ShortName, a method for regular rearrangement of objects in a room.
Different from previous methods, \ShortName learns human notions of regularity (including symmetry, alignments, uniform spacing, and stylistic and functional factors) directly from data without the need to explicitly specify a goal state.
During training, we learn from a large dataset of professionally-designed room layouts that are randomly perturbed.
During inference, we follow a Langevin Dynamics-like strategy to iteratively ``denoise'' the scene.
Quantitative results including comparisons and ablations show that our method performs well, which qualitative results confirm.

\parahead{Limitations \& Future Work}
Our method has important limitations that provide extensive opportunities for future work.
First, our method is currently limited to 2D room rearrangement and cannot perform 3D rearrangement, for instance in kitchen shelves.
However, we do incorporate 3D shape features which can be used to extend our method to 3D.
We also currently do not handle interpenetration of objects during denoising, which future work should explore.

\section*{Acknowledgements}
This work was supported by AFOSR grant FA9550-21-1-0214, NSF CloudBank, an AWS Cloud Credits award, ARL grant W911NF-21-2-0104, a Vannevar Bush Faculty Fellowship, and a gift from the Adobe Corporation. We thank Kai Wang, Daniel Ritchie, Rao Fu, and Selene Lee.

%% file: supple_content.tex
\appendix
\newpage
\section*{Supplementary Document}

\section{Overview}
This supplementary document contains extended technical details, along with qualitative and quantitative results that supplement the main document. After introducing our video results, we cover details of our network architectures and their applications (Sec~\ref{sec:architecture}). We then provide detailed explanations of our two main experiment setups: Table-Chair (Sec.~\ref{sec:table-chair}) and 3D-FRONT (Sec.~\ref{sec: 3d-front}). Next, we conduct additional analysis experiments in Sec.~\ref{sec: analysis}. Finally, we discuss failure modes (Sec.~\ref{sec: failure}) and future directions (Sec.~\ref{sec: future}).

\section{Video Results} \label{sec:video}
In order to better visualize the 3D structures of the outputs and the denoising process, we provide videos of these processes in the format of an HTML website. We highly encourage the viewers to open the file ``LEGO.html" and watch the videos for a direct view of the denoising process.

\section{Architecture Details}\label{sec:architecture}
\subsection{Input Object Attribute Encoding} 
For each object attribute $o_i=(c_i,t_i,r_i,b_i,h_i)$, we process it into an object token $\delta_i \in \mathbb{R}^{512}$ to input into the transformer. The details are as follows. 

We embed the translation $t_i$, rotation $r_i = [\cos(\theta_i), \sin(\theta_i)]^T$, and bounding box dimension $b_i$ with a sinusoidal positional encoding of 32 frequencies. The frequencies are a geometric sequence with initial term $1$ and common ratio $128^{\frac{1}{31}}$, which gives an ending term of $128$. The positional encoding is therefore 
$$PE(x) = \{\sin(128^{\frac{j}{31}}x), \cos(128^{\frac{j}{31}}x)\;|\; 0\leq j \leq 31\}\in\mathbb{R}^{64}.$$
\noindent Applying $PE$ to $t_i=[t_{i,x}, t_{i,y}]$ and $b_i=[b_{i,x}, b_{i,y}]$ gives $128$-dimensional embeddings, whereas applying it to $\theta_i$ gives a $64$-dimensional embedding. We then additionally process $PE(\theta_i)$ with a linear layer mapping to $\mathbb{R}^{128}$.

As mentioned, for object class $c_i$, we utilize a 2-layer MLP with leaky ReLU activation to process the one-hot encoding into a $128$-dimensional attribute. The above four features are concatenated to form a $512$-dimensional vector. 

We also optionally process a pose-invariant shape feature $h_i$ from ConDor \cite{sajnani2022condor}. More specifically, we pretrain ConDor on the 3D point clouds provided by 3D-FRONT \cite{fu20213d}, and extract the product of the Tensor Field Network layer output and spherical harmonics coefficients from ConDor to provide pose-invariant features in $\mathbb{R}^{1024 \times 128}$ capturing the shape of each object. Taking the mean across the $1024$ points gives a $128$-dimensional feature, which we then pass through a $2$-layer MLP with leaky ReLU activation to obtain the final shape feature in $\mathbb{R}^{128}$. We apply the shape feature $h_i$ in the {\em Grouping by Shapes} experiment to demonstrate its effectiveness.

Finally, the concatenated attributes for each object
lie in $\mathbb{R}^{640}$ and are processed through 2 linear layers with leaky ReLU activation to produce a $512$-dimensional object token for the transformer. Note that if floor plan is utilized, we modify the final layer to produce a $511$-dimensional feature, and utilize the last bit to distinguish object tokens from floor plan tokens.

\subsection{Floor Plan Encoder Architecture}
We represent each floor plan as $250$ randomly sampled contour points. We represent each point through its 2D position coordinate and the 2D normal of the line it is on. In aggregate, we represent each floor plan with a feature in $\mathbb{R}^{250 \times 4}$.

We employ a simplified PointNet \cite{qi2017pointnet} as the floor plan encoder to extract one unified floor plan feature from this representation. The encoder first processes the feature with $3$ linear layers and Leaky ReLU activation, mapping the feature dimension through $[4, 64, 64, 512]$. Then, we max pool the resulting embedding in $\mathbb{R}^{250 \times 512}$ to obtain a global floor-plan encoding in $\mathbb{R}^{512}$. We then pass this global representation through one final linear layer and append a binary bit distinguishing it from object tokens. Finally, we combine this floor plan token with the $|\tilde{X}|$ number of 512-dimensional object tokens and pass the resulting  $(|\tilde{X}|+1) \times 512$ input matrix to the transformer.

\subsection{Output Layers}
The backbone transformer outputs $(|\tilde{X}|+1) \times 512$ feature tokens. We ignore the floor plan token and use the $|\tilde{X}|$ object tokens to predict denoising transformations. We apply 2 linear layers with leaky ReLU activation to the output tokens to map to $256$ dimensions then $4$ dimensions, and finally obtain the absolute transformation predictions.

\subsection{Inference Langevin Dynamics Parameters} 

During inference, we use the Langevin Dynamics scheme to iteratively denoise a messy scene input. As mentioned, for time step $\tau$, we select $\alpha(\tau) = \alpha_0 / (1 + a_1*\tau)$ to regulate the step size and $\beta(\tau)= \beta_0 * b_1^{\lfloor \tau/b_2 \rfloor}$ to regulate the noise added at each iteration. We empirically select $a_1=0.005$ and $b_1=0.9$. For the living room, we adopt $\alpha_0=0.1$, $\beta_0=0.01$, and $b_2=10$. For bedroom, we adopt $\alpha_0=0.08$, $\beta_0=0.008$, and $b_2=8$.

We break from the iterative denoising process upon any one of two conditions: (1) if for $3$ consecutive iterations, both the predicted translation displacement vector has Frobenius norm less than $0.01$ and the predicted rotation angle displacement is less than $0.005$ radians, or (2) we have reached $1500$ iterations.

\section{Table-Chair Experiment} \label{sec:table-chair}

\subsection{Data Generation}
To analyze the regularities our model can capture, we propose three synthetic Table-Chair experiment settings, with a focus on Symmetry and Parallelism, Uniform Spacing, and Grouping by Shapes respectively.

For each of the proposed experiments, we generate clean scenes based on designed rules and take a bimodal approach at perturbations when generating clean-messy training data pair. More specifically, for half of the synthesized clean scenes, we employ a Gaussian noise kernel whose standard deviation is drawn from a zero-mean Gaussian distribution with a small standard deviation ($0.01$ for translation and $\pi/90$ for rotation angle). For the other half of the clean scenes, we employ a Gaussian noise kernel whose standard deviation is drawn from a zero-mean Gaussian distribution with a relatively larger standard deviation ($0.25$ for translation and $\pi/4$ for rotation angle). For the other training details, we follow the same paradigm as in the 3D-FRONT experiments.

\subsection{Inference Parameters}
As for the 3D-FRONT experiments, we employ the Langevin Dynamics scheme to rearrange a given perturbed Table-Chair arrangement. We empirically adjust the parameters to slightly increase the step size, accelerate the noise decay schedule, and loosen the termination condition. In particular, we select $\alpha_0=0.12$, $a_1=0.005$, $\beta_0=0.01$, $b_1=0.9$, $b_2=2$, and terminate once the predicted displacements are small enough in magnitude for $1$ iteration.

\begin{figure} [t]
    \centering
    \includegraphics[width=1.0\linewidth]{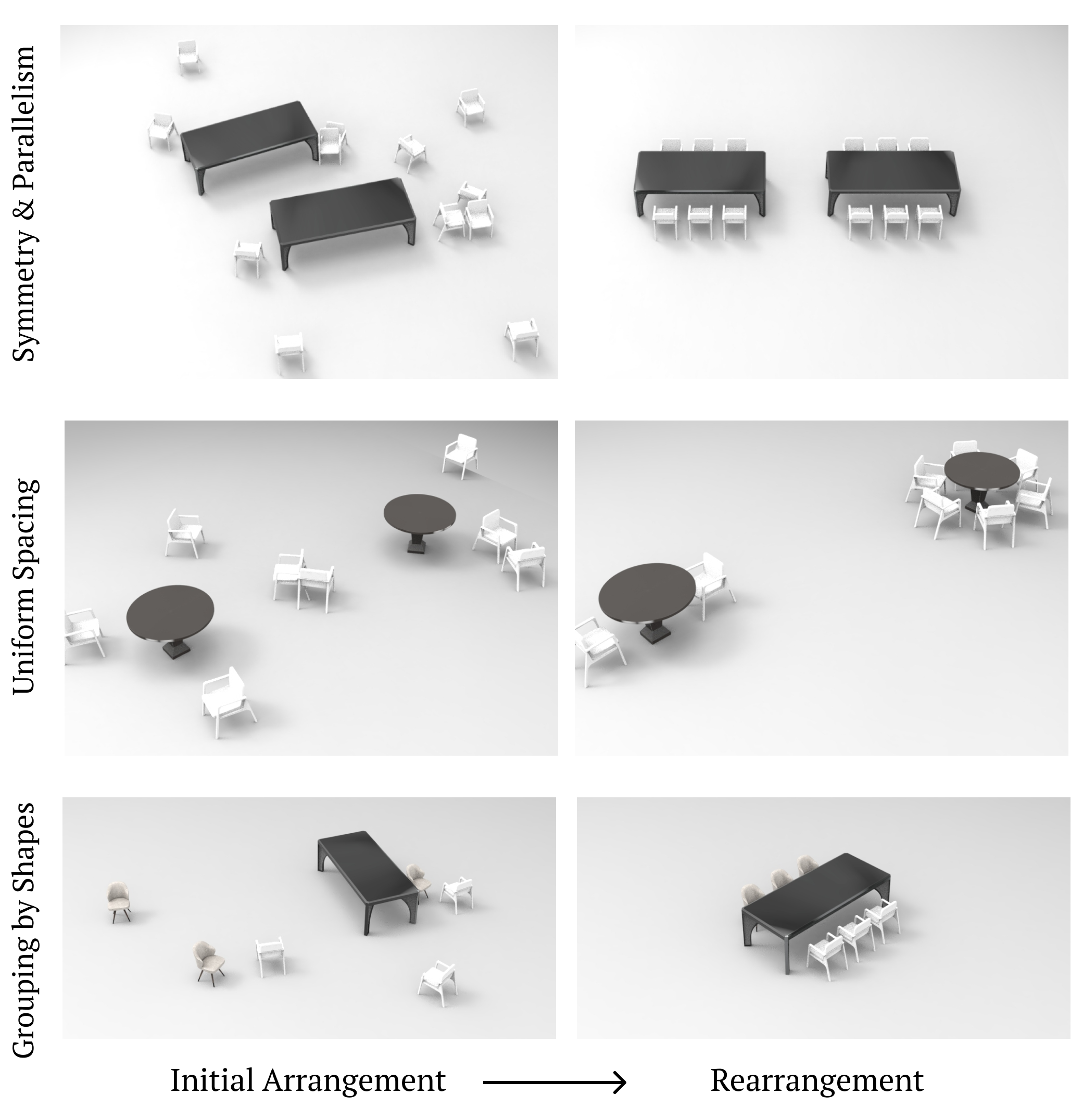}
    \caption{Regularities learning results (additional visualizations). We train our denoising network to learn three different regularities. LEGO-Net learns the complex regularity rules as demonstrated by the iterative denoising results shown on the right. Zoom in for details, especially for the shape-based grouping.}
    \label{fig: analysis_supple}
    \vspace{-3mm}
\end{figure}

\subsection{Success Rate}  \label{section:success_rate} 
As mentioned, we measure LEGO-Net's performance in each Table-Chair environment through the success rate of its rearrangement. We will now elaborate on its criteria.

\vspace*{0.1cm} \parahead{Symmetry and Parallelism}
For a rearrangement to be classified as a success, it must satisfy the following:
\begin{itemize}
    \item The mean euclidean distance of per-object movement averaged across scenes is less than $0.5$.
    \item For each chair, the angular offset between its orientation and the table-facing orientation 
    is less than $\pi/60$ radians.
    \item Given the two rearranged table positions, we compute their respective chair positions and perform Earth Mover's Distance assignment using these as target and the final predicted chair positions as source. The total distances summed across all $12$ chairs for the $2$ tables need to be less than $0.08$. Note that this metric integrates colinearity, parallelism, and symmetry, and penalizes collision.
\end{itemize}

\parahead{Uniform Spacing}
For a rearrangement in the Uniform Spacing experiment to be classified as a success, it must satisfy the first two criteria for the Symmetry and Parallelism experiment. 
For the third criteria, because the number of chairs arranged around each of the $2$ circular tables is variable, we cannot formulate the Earth Mover's Distance assignment as in the Symmetry and Parallelism experiment. Instead, to measure how well an arrangement captures the object-object relationships and regular relative positioning, we compute two other metrics. 

For each table, we compute the angular distances between each adjacent pair of its chairs and measure the variance of these distances. We designate that a successful arrangement must have an angular distance variance of less than $0.009$ radians. Additionally, given we utilize a fixed radius to generate clean chair arrangements around tables, we can measure the mean difference between the chair-to-closest-table distance and this radius. We designate that the magnitude of this difference needs to be less than $0.01$ for an arrangement to be considered successful.

\parahead{Grouping by Shapes}
Similarly, for a rearrangement in the Grouping by Shapes experiment to be classified as a success, it must satisfy the first two criteria for the Symmetry and Parallelism experiment. Additionally, we once again can compute the exact regular arrangement of chairs with respect to the table, enabling us to calculate the Earth Mover's Distance from the final predicted chair positions to the clean configuration with respect to the predicted table position. We designate that the distance summed across the $6$ chairs needs to be less than $0.05$.

Furthermore, to measure success at grouping, we require that each row must be assigned exactly $3$ chairs and that all chairs assigned to the same row must have the same shape feature.

\subsection{Additional Qualitative Results}
In Fig.~\ref{fig: analysis_supple}, we provide additional qualitative renderings for the three Table-Chair experiments.

\section{3D-FRONT Experiment}\label{sec: 3d-front}
To process the 3D-FRONT dataset \cite{fu20213d}, we closely follow the preprocessing protocol of ATISS \cite{Paschalidou2021NEURIPS}. For each scene, we extract from the given meshes and parameters the translation, rotation, class, and bounding box size for every object, and we normalize all lengths to be in $[-1, 1]$. We additionally extract accurate contours of the floor plans by running an iterative closest point algorithm \cite{besl1992method}, using the contour corner points of ATISS's binary floor plan masks as source and the relevant vertices from 3D-FRONT floor meshes as target.

\subsection{Baseline Description}
\begin{figure*}[h!]
    \centering
    \vspace{-0.1em}
    \includegraphics[width= 0.95\textwidth]{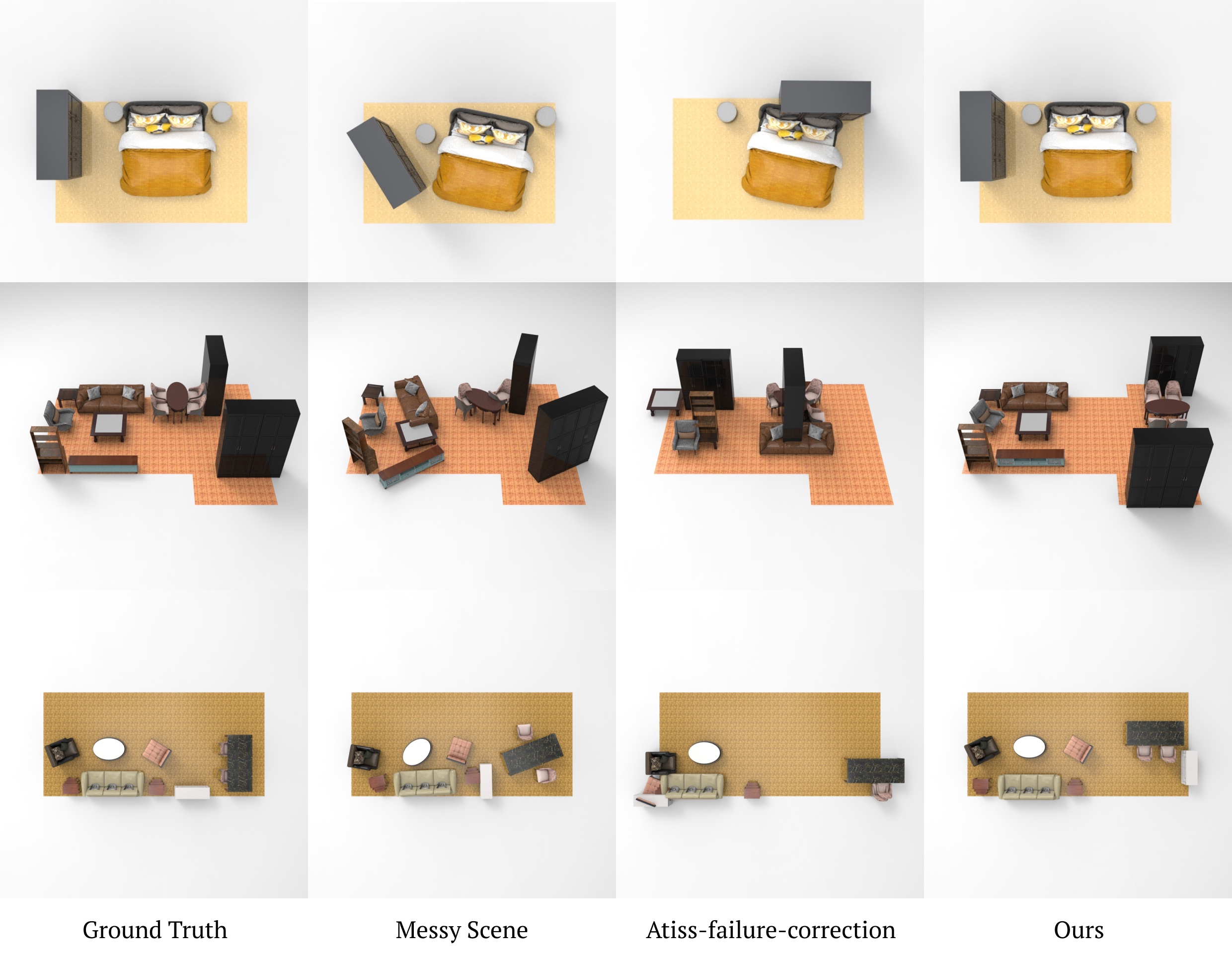}
    \caption{Qualitative comparison against ATISS-failure-correction. We show qualitative examples illustrating the difference in behaviors between our method and the closest method on the same task, ATISS {\em failure-correction}. Being a generative model, ATISS is able to compute the probability of the current object transformations and resample the ones with low probability. While ATISS has demonstrated great results in their paper when only a single object is perturbed, we observe that when {\em all} objects are perturbed, ATISS \emph{failure-correction} has a hard time fixing the scene to a reasonable regular configuration. We hypothesize that this is due to the one-prediction-at-a-time nature of ATISS -- it is difficult to find a good location to put the current object when all the other objects are perturbed. On the other hand, our method simultaneously optimizes for all the objects, avoiding such difficulty. Even for the highly challenging scene structure of the second row, our method is able to provide a high-quality layout, while ATISS \emph{failure-correction} fails to find regular rearrangement. 
    }
    \label{fig:atiss_compare}
    \vspace{-0.1em}
\end{figure*}

We compare against three variants of ATISS: {\em vanilla, labels}, and {\em failure-correction}. As described in the main text, {\em vanilla} is the original ATISS approach that generates a scene from scratch given the floor plan. ATISS {\em labels} is given the floor plan, as well as a set of furniture labels and the transformations and sizes of the labeled objects.
ATISS {\em failure-correction} is proposed as an application to the probabilistic generative modeling of ATISS. It identifies which object is likely to be a failure and resamples that object given all the other objects. While the original paper only showed the technique to work when a single object is perturbed, we find it reasonable to extend the algorithm to multi-object perturbation cases. Specifically, we provide a scene with all objects perturbed (same input as LEGO-Net) and iteratively resample the lowest-probable object. We stop the iteration when it reaches 1,000 times or when the minimum probability is higher than a manually set threshold. We note that while {\em failure-correction} did not perform as expected when all of the objects are perturbed, as shown in Fig.~\ref{fig:atiss_compare}, it is the closest baseline we could find in the literature that performs data-driven denoising of a scene.

For the comparisons, we use the official training and testing code provided by the authors of ATISS without modifications. For ATISS {\em failure-correction}, we add a for-loop and stopping criteria on top of their implementation, and maintain the scales of objects as fixed to be coherent with the rearrangement task.

\subsection{Metric Description} 

For FID and KID computation, we first generate the same number of scenes as in the test dataset and randomly select 500 from the generated scenes. We then randomly select 500 real scenes from the 3D-FRONT dataset to compare against. For both FID and KID, we repeat the metric computation 5 times and report the average. Note that for computing the FID scores for ATISS, we used the officially provided code and followed their exact evaluation procedure, but we failed to reproduce their numbers. Hence, we use our own way of computing the metrics and report ours.

We note that the Frechet Inception Distance (FID) \cite{heusel2017gans} is known to present significant positive bias when the number of images is small (e.g., $\leq$ 2000). In our case, we're dealing with an even smaller number of test images. Therefore, to compute a metric that is less biased in the small-data regime, we adopt the Kernel Inception Distance \cite{binkowski2018demystifying}, which is known to address the bias problem and present small variance even at a few hundred samples.

\paragraph{Distance Traveled}
As discussed in the main text, we aim at rearranging the messy scenes while retaining the flavor of the original scenes. Practically, cleaning a room should move objects as minimally as possible while realizing regularities. Therefore, we measure and report the mean of the average distance traveled for scenes. More specifically, for each scene, we compute the average Euclidean distance between the corresponding objects in the initial and final states, and take the mean across scenes.

Note that for ATISS \emph{vanilla}, this metric is not applicable as the method randomly places objects into the scene. For ATISS {\em failure-correction}, we calculate this metric by computing the distance between the initial position of each object and its final position after applying the algorithm.

\paragraph{EMD to GT}
In order to measure how accurately our method recovers the original scene configuration, we measure the Earth Mover's Distance between the final and the ground truth scene states. Note that computing the difference between the denoising prediction and the ground truth is widely used in the image-denoising literature, using such metrics as PSNR or SSIM \cite{hore2010image}. ATISS {\em vanilla} and {\em labels} do not receive the messy scene as input, so this metric is not applicable to them. On the other hand, ATISS {\em failure-correction} directly fixes the input scene, thus we may measure how accurately it recovers the original clean scene. Finally, we note that the EMD to GT metric becomes highly noisy and irrelevant when the noise added to perturb the clean scenes becomes too high, as then, there is scarsely any locational information left in the messy inputs.

\subsection{Additional Qualitative Results}
\begin{figure*}
    \centering
    \includegraphics[width= 0.91\textwidth]{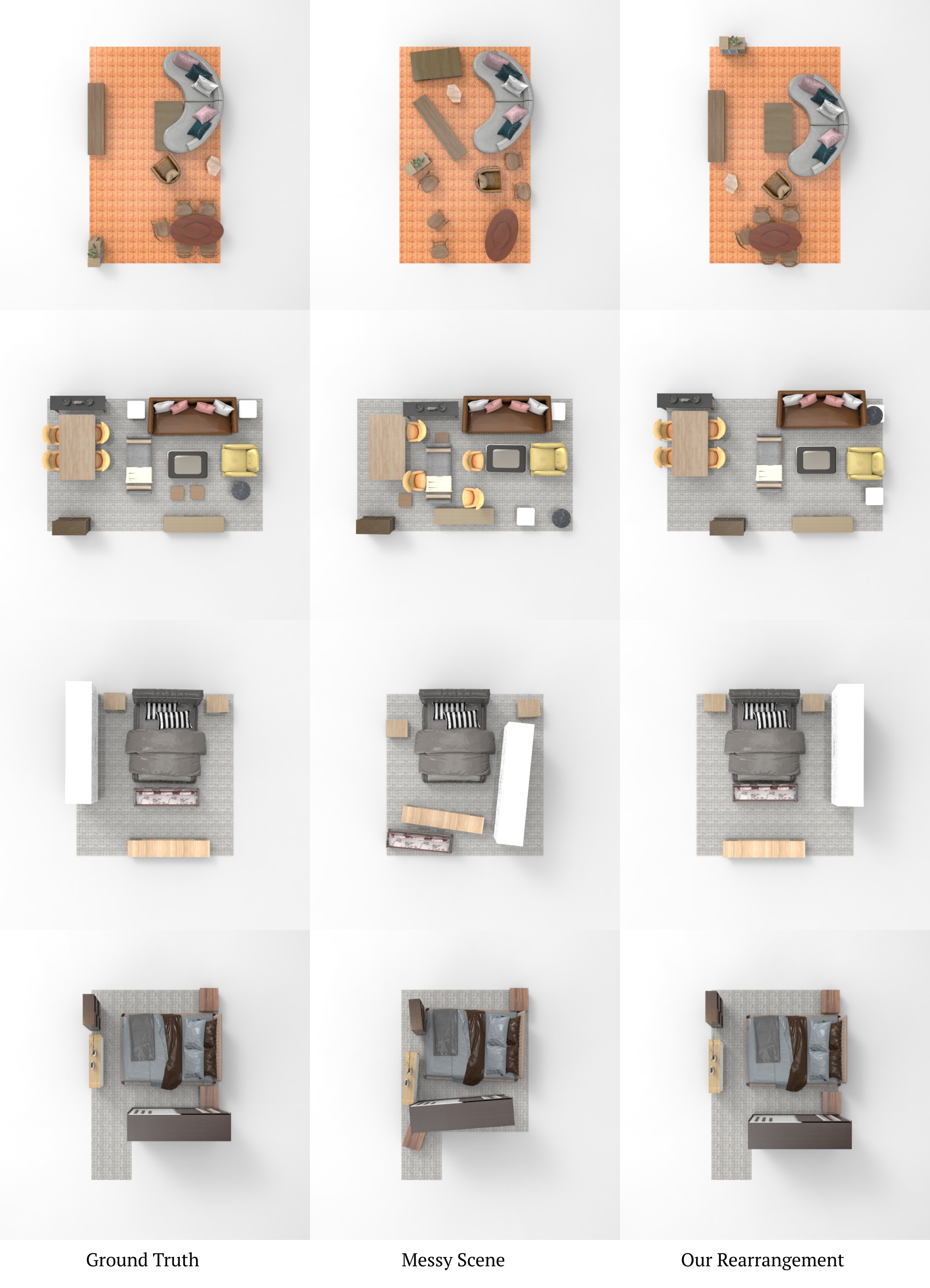}
    \caption{Additional rearrangement results of LEGO-Net on the 3D-FRONT dataset.
    }\label{fig: additional}
\end{figure*}

We provide additional qualitative results on the 3D-FRONT dataset. We show more comparisons against the closest method on the scene rearrangement task, ATISS {\em failure-correction}, in Fig.~\ref{fig:atiss_compare}, and more results of our method in Fig.~\ref{fig: additional}.

\section{Analysis (Continued)} \label{sec: analysis}

\begin{table}
\centering
\begin{tabular}{ c c c }
\toprule
 & Bedroom $\uparrow$
 & Living Room $\uparrow$
  \\ \hline
PointNet W/O Noise & \textbf{84.4}\% & \textbf{54.4}\%     \\
PointNet W/ Noise & 84.2\% & 54.2\%   \\ 
ResNet W/O Noise & 83.8\% & 54\%  \\ 
ResNet W/ Noise  & 83.6\% & 54.2\%    \\
\bottomrule
\end{tabular}
\caption{Percentage of denoised scenes with 90\% of its furniture within the floor plan boundaries. We train and test our method using different floor plan encoding architectures (PointNet vs. ResNet), and measure the percentage of the denoised scenes where most furniture respect the room boundaries.}
\label{table:floorplan_encoder}
\end{table}

\subsection{Enforcing Floor Plan Constraints}
In this section, we analyze our choice of floor plan encoder. As described in the main text, we extract points on the boundary of the binary floor plan mask, and process them with a PointNet \cite{qi2017pointnet} to obtain a unified feature vector describing the floor plan. We note that, in ATISS \cite{Paschalidou2021NEURIPS}, a 2D convolutional network with residual connections (ResNet) was used to process the floor plans. Here, we conduct an experiment to justify our use of PointNet architecture. As a baseline, we use the ResNet architecture from ATISS, but augment the input floor plan with two additional channels corresponding to the {\em xy} coordinate for each pixel center, which is known to provide ``spatial awareness" to the 2D CNN (in CoordConv~\cite{liu2018intriguing}). We expect this variant of ResNet to work at least as well as the vanilla ResNet with binary mask input used in ATISS.

To compare the two methods of floor plan encoding, we train two variants of the model, using ResNet or PointNet floor plan encoding architecture. To test the effectiveness of the two methods, we measure how often furniture is moved outside the floor boundaries. Specifically, a scene rearrangement is considered successful when 90\% of objects are placed inside the floor boundary within 4\% of its length margin. While respecting the floor boundaries does not necessarily lead to high-quality, regular scenes, we empirically find this metric as a reasonable proxy. As can be seen from the numerical results of Tab.~\ref{table:floorplan_encoder}, the use of PointNet outperforms that of ResNet by a slight margin. However, we note that using PointNet is significantly faster, having almost no computational overhead for operating on the sampled 250 boundary points. We, therefore, choose to use the simpler but similar-performing PointNet to encode the scene floor plans.

\begin{table}
\centering
\begin{tabular}{ c c c c }
\toprule
 & \vtop{\hbox{\strut Distance }\hbox{\strut Moved $\downarrow$}} & \vtop{\hbox{\strut Direction}\hbox{\strut Offset $\downarrow$}}& \vtop{\hbox{\strut EMD}\hbox{\strut to GT $\downarrow$}} \\ \hline
Absolute & 2.98e-1 & 1.10e-3 & \textbf{5.16e-2} \\
Relative & \textbf{2.47e-1} & \textbf{4.21e-4} & 1.58e-1 \\ 
Relative Translation \\\& Absolute Rotation & 2.73e-1 & 5.18e-4 &
2.83e-1 \\ 
\bottomrule
\end{tabular}
\caption{Mean statistics from denoising results using the gradient with noise strategy for the Uniform \& Parallelism Table-Chair experiment. The relative prediction model slightly outperforms the absolute prediction model in terms of distance moved and angular orientation offset, but it performs much worse in terms of EMD to GT as although it maps objects to the right general region, it does not place them as precisely as the absolute model.}
\label{table:tablechair_stats}
\end{table}

\subsection{Relative vs Absolute}
2D diffusion models perform better at predicting the noise than the un-noised images \cite{ho2020denoising}. 
However, for our setting, we observed that the absolute prediction models generally outperform their relative counterparts in the Table-Chair environment. We trained $3$ variants of the same architecture network for the Uniform \& Parallelism Table-Chair setting: (i) absolute translation and rotation prediction, (ii) relative translation and rotation prediction, and (iii) relative translation and absolute rotation prediction. 

Following the criteria in \ref{section:success_rate}, both variants (ii) and (iii) surprisingly report success rates of $0\%$. Upon further investigation (shown in Tab.~\ref{table:tablechair_stats}), variants (ii) and (iii) perform comparably, if not better, at limiting the distance of movement and orienting chairs to face the tables, but they significantly underperform (i) in terms of Earth Mover's Distance to Ground Truth chair positions with respect to the predicted table position. With the same denoising parameters, the relative variants consistently fail to place the chairs as precisely as the absolute variant. 
The superior performance of the absolute variant may partly be explained by the fact that this setting has a relatively limited space of possible regular arrangements.
The observed deficiency of relative predictions may diminish as the complexity of the scene increases.

We believe that relative transformation prediction is an important future direction to explore, as they offer translation invariance, which is particularly valuable for large-scale scenes. Currently, the position and orientation information in our input object attributes are global, and thus our system is not translationally invariant. An interesting direction for future investigation is to explore a sliding-window-style input processing to ensure translation invariance and to apply positional encoding to the output transformations.

\begin{figure*}[h!]
    \centering
    \vspace{-0.1em}
    \includegraphics[width= 0.95\textwidth]{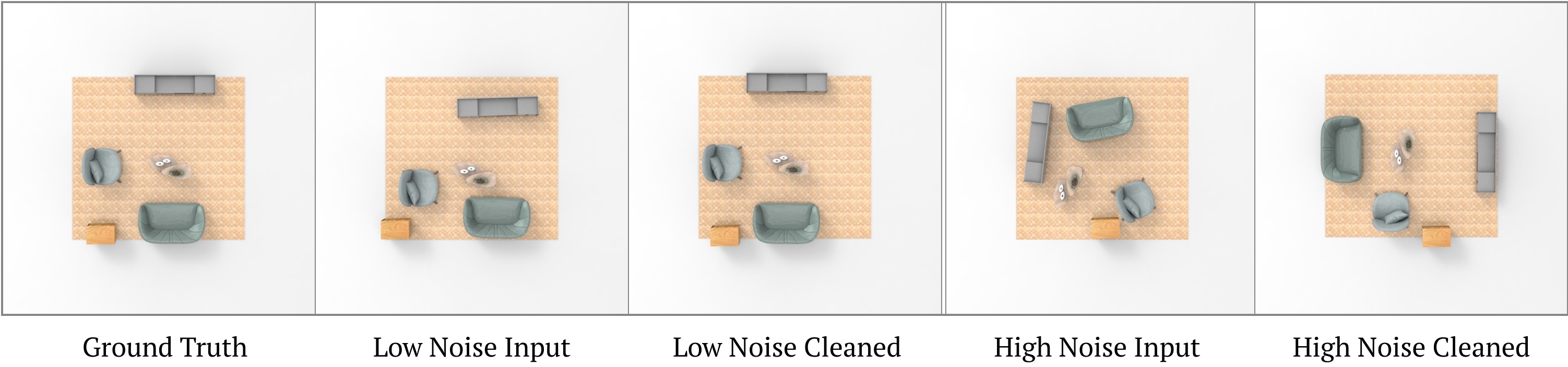}
    \caption{LEGO-Net denoising results on different noise levels (additional visualization to Fig.7 of the main text). When the perturbation added to the scene is low, LEGO-Net is able to closely reconstruct the clean version of the scene. In contrast, when the noise level is high, our denoising process finds a different realization of a regular scene, behaving more like an unconditional model. We provide numerical evidence for this phenomenon in Tab.~\ref{fig:noiselevel_graph}.    
    }
    \label{fig:noiselevel2}
    \vspace{-0.1em}
\end{figure*}

\begin{figure} [h!]
    \vspace{-0.2cm}
    \includegraphics[width= 0.48\textwidth]{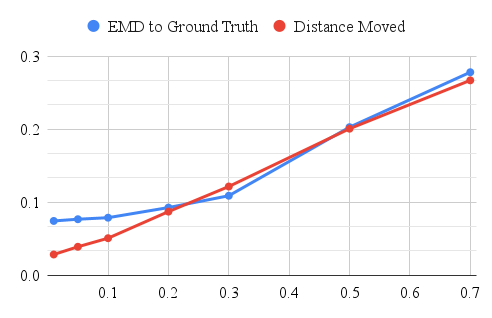}
    \caption{LEGO-Net denoising results on different noise levels (supplementing Fig.8 of the main text). The horizontal axis specifies the standard deviation of the zero-mean Gaussian distribution from which the (translation) noise level is drawn, and the vertical axis specifies distance in scenes normalized to $[-1, 1]$. We compute mean statistics across 100 scenes at $[0.01, 0.05, 0.1, 0.2, 0.3, 0.5, 0.7]$ for the standard deviation of the translation noise level distribution and at $[\pi/90, \pi/24, \pi/12, \pi/4, \pi/3, \pi/2, \pi]$ for that of rotation angle. 
    }
    \label{fig:noiselevel_graph}
    \vspace{-0.1em}
\end{figure}

\begin{figure*} [h!]
    \centering
    \vspace{-0.1em}
    \includegraphics[width= 0.99\linewidth]{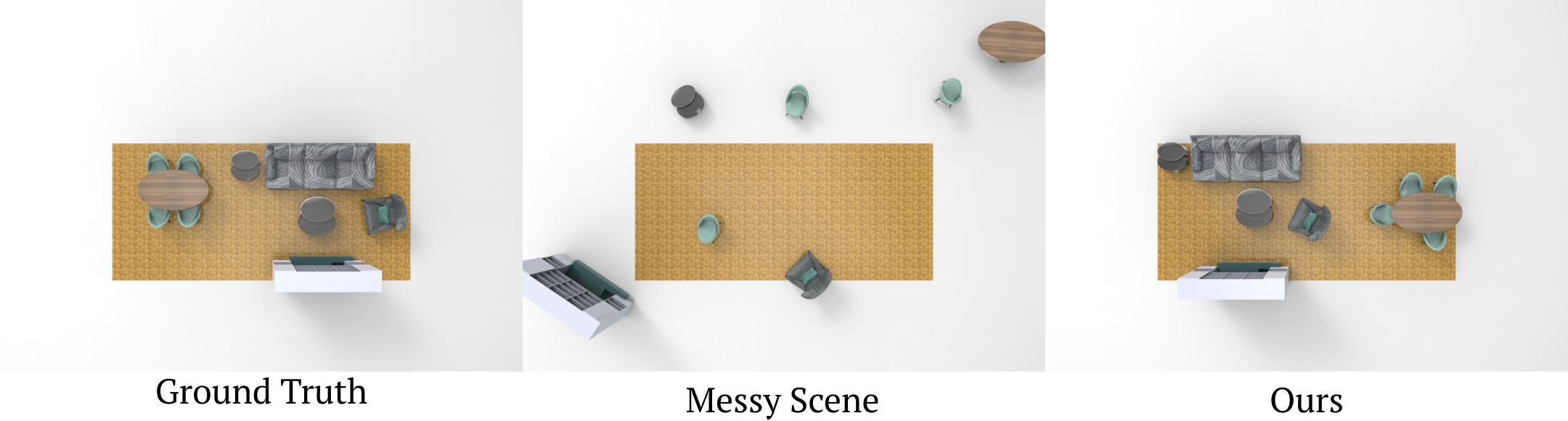}
    \caption{Extreme case rearrangement. When the amount of noise added is exceedingly high, the original scene structure is gone. Running our denoising algorithm then amounts to unconditional sampling and leads to an arrangement significantly different from the ground truth.
    }
    \label{fig:extreme}
    \vspace{-0.1em}
\end{figure*}

\subsection{Distance to Ground Truth and Distance Moved vs. Noise}
Since the task of rearrangement values affinity to the starting configuration of objects, one question of interest is how closely we recover the original clean arrangement when given a perturbed version of it, versus another possibly equally valid, clean arrangement. Its correlation with the level of noise added is intuitive--we expect that when the perturbation is low, we more closely reconstruct the original scen with a low distance moved whereas when the perturbation is high, our model may choose a regular arrangement different from that of the original scene in an effort to minimize the distance moved (see Fig.~\ref{fig:noiselevel2}). This is indeed what we have observed numerically, as shown in Fig \ref{fig:noiselevel_graph}.

Note that with a low degree of noise, our model performs the task of rearrangement, but as the degree of noise increases, our model gradually transitions to the task of arrangement. In the extreme case where we give \ShortName a scene with objects outside of the floor plan, \ShortName is able to perform scene arrangement from scratch (see Fig.~\ref{fig:extreme}).

The open-endedness in the definition of regularity is one of the reasons why this task is both challenging and interesting. The interpolation-like behavior of \ShortName conditioned on the degree of noise signifies the learnable relationship between rearrangement and synthesis, and it demonstrates that a diffusion-like approach holds promising potential at such open tasks.

\begin{figure*} [h!]
    \centering
    \includegraphics[width= 0.95\textwidth]{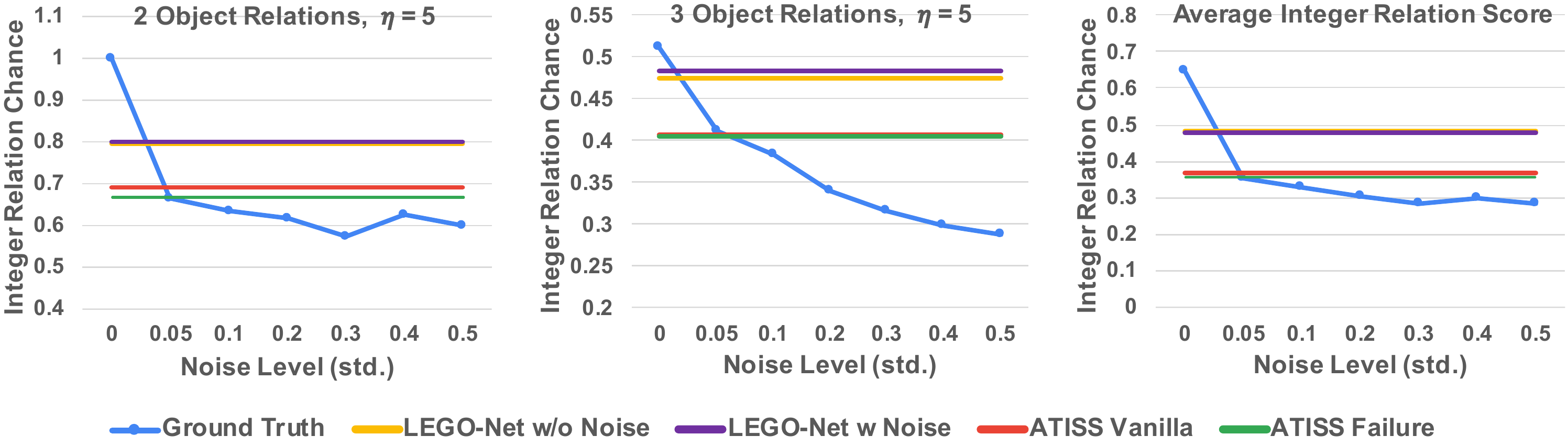}
    \caption{
    We measure the chance of finding integer relations among the coordinates of 2 (left) and 3 (middle) objects within a living room scene for $\eta=5$. We aggregate these two settings with the two presented in the main text to produce an overall average integer relation score (right) for general regularity. We normalize all raw chance measures by the maximal chance across the four settings, making all numerical values directly comparable. Note that \ShortName outperforms the ATISS variants.
    }
    \label{fig:integer_relations_moregeneral}
\end{figure*}

\subsection{Integer Relations} 
The task of evaluating regularity in an object arrangement is itself an interesting research problem because, in general, multiple regular solutions are possible. To evaluate and quantify the notion of ``regularity'' in object arrangements, we propose using number-theoretic machinery for detecting and evaluating sparse linear integer relations among object coordinates. That is, given coordinates $t_i$'s for $n$ objects, we seek to find integral coefficients $a_i$'s such that: 
\begin{equation}\label{eq:integer2}
    a_1t_1+...+a_nt_n = 0, \quad 0<|a_i|<\eta\,, \forall a_i,
\end{equation} 

For simplicity, we consider each dimension of the cooridnates separately, i.e., $t_i\in \mathbb{R}.$ Additionally, in practice, we introduce an additional parameter $\epsilon$ to control the precision of the solutions found. In particular, we seek to find relationships such that $|a_1t_1+...+a_nt_n| < \epsilon$. For our evaluation, we fix $\epsilon=0.01$ to focus on the near-perfect relations while allowing some leeway for insignificant offsets. To efficiently find these integer solutions, we employ the PSLQ algorithm ~\cite{Ferguson1991}.

When the subset size $n$ and maximum coefficient magnitude constraint $\eta$ are small, the integer relations can be intuitively understood (e.g. representing co-linearity, symmetry, uniform spacing among few objects). To capture more complex and more general notions of regularity, we increase $n$ and $\eta$. Doing so preicipitates two challenges. First, the number of possible subsets of size $n$ for each scene increases rapidly as $n$ increases, compromising efficiency. Secondly, experimentally running the algorithm on pure noise shows that with looser constraints, we may find many trivial relations that do not appear to correspond to high-level `cleanness'. To counter these, we introduce two filtering mechanisms to increase the subset sampling efficiency and to filter out the insignificant relations.

We observe that regualrities of interest to us mostly occur among objects in close proximity to one another, such as tables and chairs. Therefore, for $n>2$, instead of sampling from all possible subsets of the objects in the scene, we iterate through each object and sample from the object's positional neighborhood. For $n=3$, for each object, we sample $2$ from the closest $4$ neighboring objects to form $\{t_1, t_2, t_3\}$. This greatly improves sampling efficiency, and it also helps eliminate irrelevant candidates as integer relations satisfied by objects in vicinity to one another are more likely to be meaningful for the purpose of our evalutions.

Additionally, to filter out relations that may have been satisfied by numerical coincidence, we require all relations to be translation-invariant. Specifically, for each subset, we sample a noise $\mu$ from uniform distribution $U(-1,1)$ and apply the PSLQ algorithm to $\{t_1+\mu, ..., t_n+\mu\}$. We repeat the process $10$ times and only deem a subset to have a valid relation if the algorithm succeeds for all $10$ times. This helps the metric to focus on regularities with respect to the relative positions instead of the absolute positions of objects.

In Fig \ref{fig:integer_relations_moregeneral}, we demonstrate that with these two filtering mechanisms, our metric is still meaningful for $\eta=5$ and potentially larger parameters, which would be useful for measuring wider ranges of regularities. We also show that averaging all the integer relation metrics across the various settings suggest more general notions of regularity, for which \ShortName outperforms the ATISS variants.

\begin{figure} [h]
    \centering
    \includegraphics[width=1.0\linewidth]{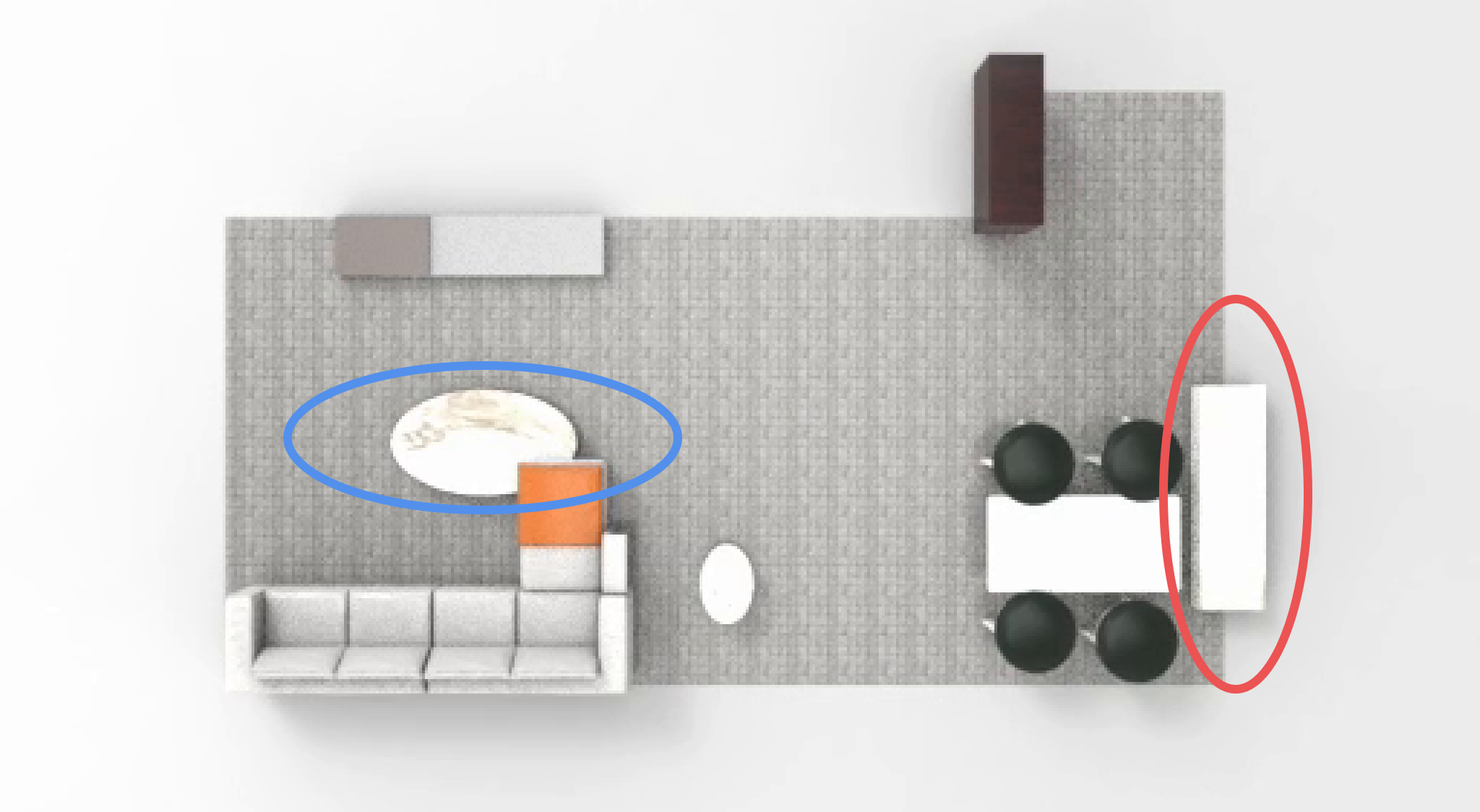}
    \caption{Failure modes. While LEGO-Net generates high-quality scenes, it often exhibits the two failure modes of placing objects outside of the floor plan boundaries (red ellipse) and inducing penetration between objects (blue ellipse). We leave post-processing steps to resolve these issues as future work.}
    \label{fig: failure}
    \vspace{-3mm}
\end{figure}

\section{Failure Modes} \label{sec: failure}
\begin{figure*}[b]
    \centering
    \includegraphics[width= 1\textwidth]{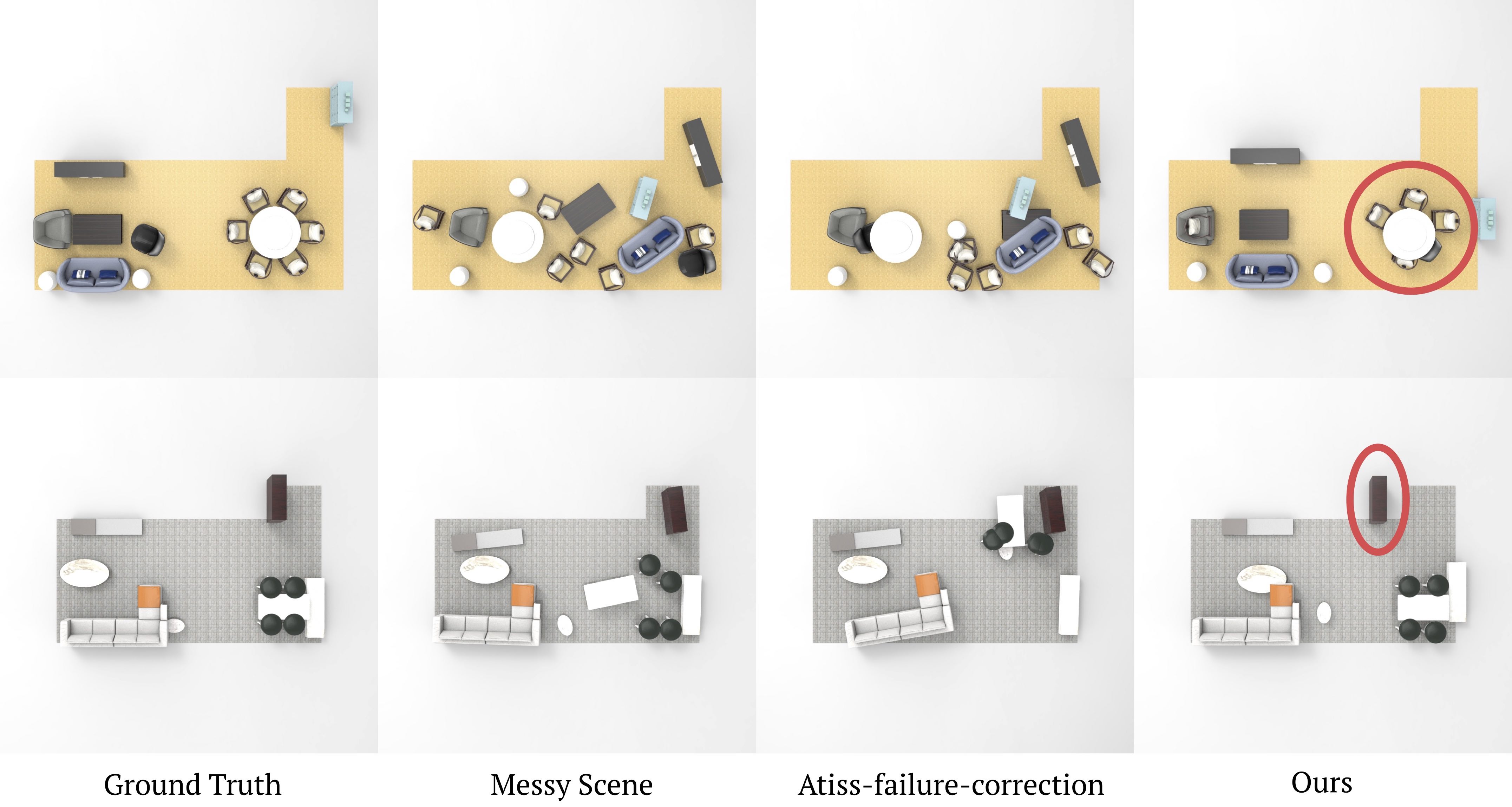}
    \caption{Additional demonstrations of failure modes of our method. In the top row, note that two chairs are missing due to perfect collision of their position predictions with those of the other chairs. In the bottom row, the cabinet is placed outside of the floor boundaries.
    }\label{fig: morefailure}
\end{figure*}

While LEGO-Net generates unprecedented-quality indoor scenes through the iterative denoising process, we notice that it often suffers from objects going out of the floor boundaries and objects penetrating each other. These failure modes are illustrated in Fig.~\ref{fig: failure} and ~\ref{fig: morefailure}. In fact, in Tab~\ref{table:floorplan_encoder}, we measure that around half of living room realizations have at least one object outside of the boundaries.

We propose two possible remedies for these related issues. First, we could apply post-processing steps to physically resolve the two problems. That is, we optimize the locations of the objects within the scene such that penetrations and out-of-boundary issues are resolved with minimal required movement. We believe a possible formulation would involve a signed distance function (SDF), with which it is easy to compute the gradient to minimize the penetrations. When a point lies on the negative territory of another shape's SDF, we can optimize the location of that point out towards the SDF's gradient directions.

Secondly, one can consider richer encoding of the floor plan. One possibility is to encode each line segment of the floor plan separately as a token. This will essentially treat each line segment as an object in the scene and could enforce stricter constraints on the boundaries. At a glance, this strategy might increase the computational cost significantly, due to the quadratic nature of Transformer time complexity. However, one could consider limiting the communication between the line segment tokens to prevent quadratic scaling of the complexity. 

We leave these two potential remedies for our failure modes as future work.

\section{Future Work} \label{sec: future}
In this work, we introduced LEGO-Net, an iterative-denoising-based method for tackling scene rearrangement task, which is relatively understudied compared to the scene synthesis task. We show through extensive experiments that our method is able to capture regularities of complex scenes, generating high-quality object rearrangements that could not be achieved by existing approaches to date. 

However, LEGO-Net in its current form only operates on a 2D plane for the rearrangement. Extending our work to operate on the $SE(3)$ transformation space would make it more applicable to real-world scenes. A significant barrier to achieving 3D rearrangement is the lack of data. We notice that most of the indoor scene arrangement datasets deal with objects laid on the floor plan, which could limit the progress of studying $SE(3)$ scene arrangements. Designing and collecting such a dataset, e.g., small objects on top of one another is an interesting future direction.

Moreover, the trajectories generated during the denoising process of LEGO-Net are not meant to respect physical constraints, e.g., penetrations and collisions. We find that enforcing the physical constraints during the denoising steps could significantly limit the space of possible scene rearrangement. Currently, if one wants to move objects in a scene according to the initial and final states of our algorithm, one needs to run a motion planning algorithm. Extending our work to output motion plans, along with the final states, is worth pursuing.

Finally, LEGO-Net has only been shown to work well on relatively small room-scale scenes. Extending our work to operate on larger-scale scenes such as warehouses might require changes to some of our architecture choices, including strengthening translational invariance. Exploring such strategies remains an understudied challenge, which we continue to explore.

